\documentclass{ifm-tr}

\usepackage{amsmath,amsfonts,bm}

\def\eqref#1{equation~\ref{#1}}

\def\1{\bm{1}}

\DeclareMathAlphabet{\mathsfit}{\encodingdefault}{\sfdefault}{m}{sl}
\SetMathAlphabet{\mathsfit}{bold}{\encodingdefault}{\sfdefault}{bx}{n}

\usepackage{amsmath,amsfonts,amssymb,mathtools,bm}  
\usepackage{thmtools}          
\usepackage{algorithm}         
\usepackage{algpseudocode}     
\usepackage{enumitem}          
\usepackage{siunitx}           
\usepackage{tikz}              

\usepackage{hyperref}       
\usepackage{url}            

\usepackage{xurl}          
\usepackage{booktabs}       
\usepackage{amsfonts}       
\usepackage{nicefrac}       
\usepackage{microtype}      
\usepackage{graphicx}
\usepackage{subcaption} 
\usepackage{xspace}
\usepackage[frozencache,cachedir=.]{minted}
\usepackage{longtable}
\usepackage{adjustbox}

\usepackage{nicematrix}
\usepackage{multirow}
\usepackage{makecell}
\usepackage{soul}  
\usepackage{wrapfig}
\usepackage{enumitem}
\usepackage{tabularx}

\usepackage{lipsum}
\usepackage{tcolorbox}
\usepackage{fancyvrb}
\tcbuselibrary{breakable}
\usepackage{fontawesome5} 
\usepackage{pdflscape} 

\usepackage{colortbl}
\usepackage[normalem]{ulem}
\usepackage{pgfplots}
\usepackage[table]{xcolor}
\pgfplotsset{compat=1.17}

\definecolor{colbg}{HTML}{F2F2F2} 
\definecolor{dimrule}{HTML}{CCCCCC} 

\usepackage{microtype} 
\usepackage{listings}
\usepackage{placeins}
\usepackage{xspace}

\definecolor{cellHighlight}{HTML}{dbefff}
\definecolor{promptbg}{gray}{0.95}

\newcommand\rurl[1]{%
    \href{https://#1}{\nolinkurl{#1}}%
}

\newcommand\grurl[1]{%
    \href{https://github.com/#1}{\nolinkurl{#1}}%
}

\newcommand\hrurl[1]{%
    \href{https://huggingface.co/#1}{\nolinkurl{#1}}%
}

\newcommand\wrurl[1]{%
    \href{https://wandb.ai/#1}{\nolinkurl{#1}}%
}

\newtcolorbox{AIbox}[2][]{aibox,title=#2,#1}

\definecolor{blue-primary}{HTML}{0081FB}
\definecolor{blue-light}{HTML}{D4ECFF}
\newtcolorbox{keytakeaways}[1][]{
    enhanced,
    colback=blue-light, 
    colframe=blue-primary,
    colbacktitle=blue-primary,
    coltitle=white,
    fonttitle=\bfseries\large\color{white},
    title={\faLightbulb\, Key Takeaways},
    after=\par\vspace{0.5em},
    arc=2.5pt,
    boxrule=0.6pt,
    left=9pt,
    right=9pt,
    top=8pt,
    bottom=8pt,
    drop fuzzy shadow, 
    #1
}
\newcounter{takeawaycounter}
\newcommand{\takeaway}[1]{%
  \par\noindent%
  \begin{minipage}{\linewidth}%
    \refstepcounter{takeawaycounter}%
    {\setlength{\fboxsep}{2pt}%
     \colorbox{cyan!10}{\textbf{Takeaway \thetakeawaycounter.}}}~\textit{#1}%
  \end{minipage}\par%
}

\title{\bf EMO: Frustratingly Easy Progressive Training of\\[0.25em] Extendable MoE}

\author[†]{Linghao Jin}\authorsep{,}
\author[†]{Chufan Shi}\authorsep{,}
\author[†]{Huijuan Wang}\authorsep{,}
\author[†]{Nuan Wen}\authorsep{,}
\author[o]{Zhengzhong Liu}\authorsep{,}
\author[o]{Eric Xing}\authorsep{,} 
\author[†,o]{Xuezhe Ma}\authorsep{}

\affiliation[†]{USC-ISI}\affiliationsep{}
\affiliation[o]{MBZUAI-IFM}

\reportnumber{2026-05-12}

\abstract{
  Sparse Mixture-of-Experts (MoE) models offer a powerful way to scale model size without increasing compute, as per-token FLOPs depend only on $k$ active experts rather than the total pool of $E$ experts.
  Yet, this asymmetry creates an MoE efficiency paradox in practice: adding more experts balloons memory and communication costs, making actual training inefficient. 
  We argue that this bottleneck arises in part because current MoE training allocates too many experts from the beginning, even though early-stage data may not fully utilize such capacity.
  Motivated by this, we propose \textbf{EMO}, a simple progressive training framework that treats MoE capacity as \emph{expandable memory} and grows the expert pool over the course of training. 
  EMO explicitly models sparsity in scaling law to derive stage-wise compute-optimal token budgets for progressive expansion.
  Empirical results show that EMO matches the performance of a fixed-expert setup in large-scale experiments while improving wall-clock efficiency. It offers a surprisingly simple yet effective path to scalable MoE training, preserving the benefits of large expert pools while reducing both training time and GPU cost.
}

\begin{document}
\maketitle

\renewcommand{\thefootnote}{\fnsymbol{footnote}}
\footnotetext[1]{Correspondence to: \texttt{linghaoj@usc.edu}}
\renewcommand{\thefootnote}{\arabic{footnote}}

\section{Introduction}

\begin{wrapfigure}{r}{0.5\textwidth}
    \centering
    \vspace{-8mm}
    \includegraphics[width=0.5\textwidth]{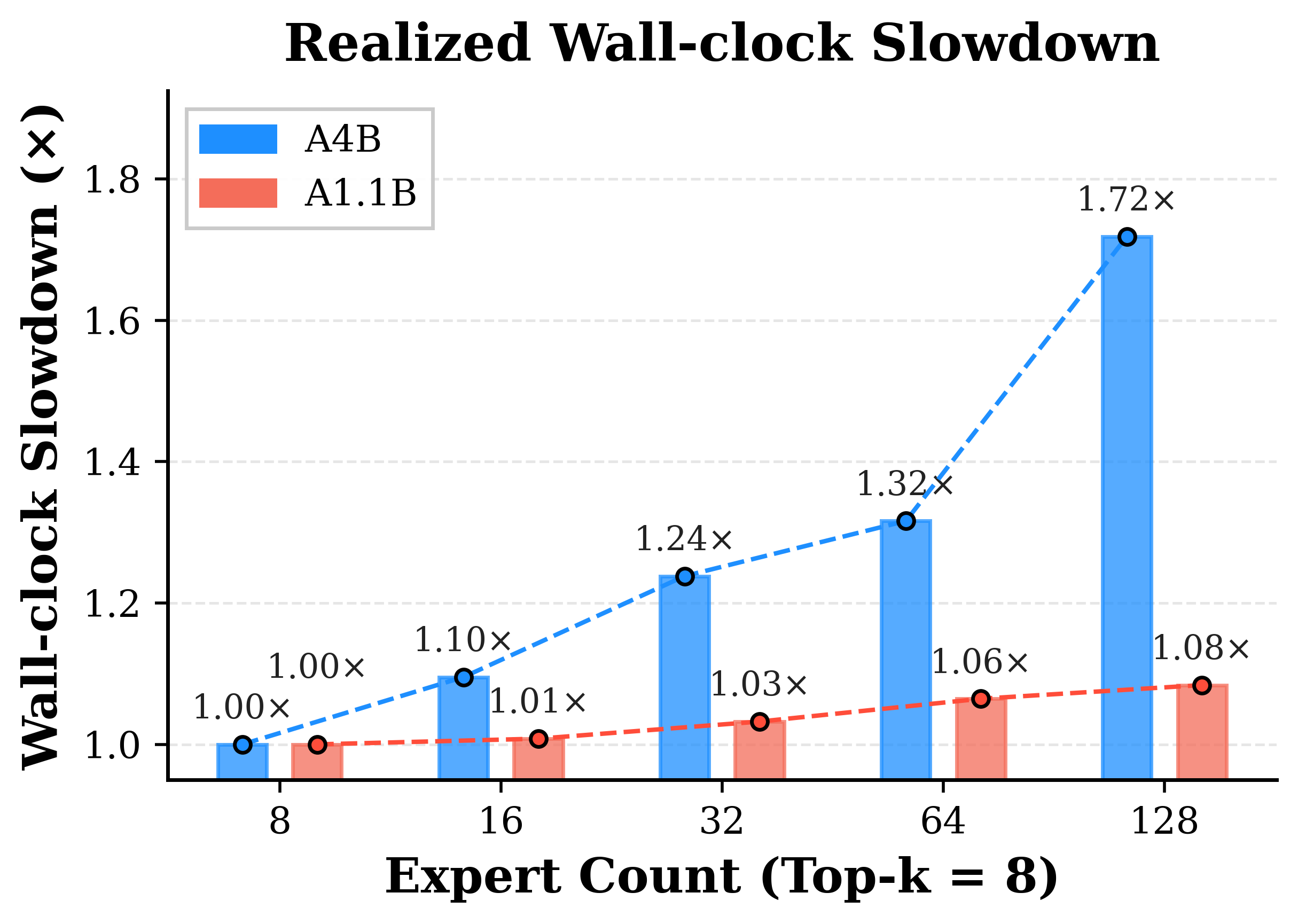}
    \caption{Increasing expert count $E$ with fixed \emph{top-k} activated experts substantially slows down training, especially at larger scales. \textit{A4B} denotes 4B activated parameters (out of 36B total at $E{=}128$); \textit{A1.1B} denotes 1.1B activated parameters (out of 9.6B at $E{=}128$). All experiments are conducted on 4 nodes of 8$\times$H200 GPUs.}
    \vspace{-5mm}
    \label{fig:intro}
\end{wrapfigure} 

Sparse Mixture-of-Experts (MoE) architectures decouple model capacity from per-token computation, making it possible to scale parameters without a proportional increase in floating-point operations (FLOPs)~\citep{shazeer2017outrageously, lepikhin2021gshard, fedus2022switch}. This decoupling has powered some of the most capable language models to date, including 
GLaM~\citep{du2022glam}, Mixtral~\citep{jiang2024mixtral},  Kimi K2~\citep{team2025kimi} and DeepSeek-V4~\citep{deepseekv4}. 
However, wall-clock time has not been fully decoupled with $E$. As $E$ grows, all-to-all communication, optimizer-state memory, and small per-expert general matrix multiplications (GEMMs) all grow with it~\citep{lepikhin2021gshard, gale2023megablocks, rajbhandari2022deepspeed, yan2026scalable}, so realized step time tracks $E$ even when FLOPs do not. \autoref{fig:intro} quantifies this gap on our setup: with fixed 4B activated parameters, increasing $E$ from 8 to 128 leaves theoretical FLOPs unchanged but inflates wall-clock step time by $1.72\times$, a gap that widens further as activated size scales up. Thus, the theoretical efficiency of MoE is largely a FLOPs-level promise, its realized training cost still grow substantially with the total number of experts.
This raises a question that remains under-appreicated in the pretraining setting: does training actually need all E experts from the start?

A useful lens is to view each expert as a unit of \emph{parametric memory}: a specialized subnetwork that stores and retrieves patterns relevant to a subset of the data~\citep{shazeer2017outrageously, roller2021hash, bengio2013estimating}. 
When training data is limited, a large expert pool is both unnecessary and inefficient---it amplifies communication costs and memory, and provides capacity that cannot yet be utilized; as data scales, additional experts (memory) become beneficial.
Recent MoE scaling laws are consistent with this view: the compute-optimal expert count grows with the token budget~\citep{ludziejewski2025joint, clark2022unified}.
This suggests a simple principle:
\begin{center}
\emph{MoE capacity should grow progressively with data as an expandable memory.} 
\end{center}

Building on this principle, we propose \textbf{EMO} (\textbf{E}xtendable \textbf{M}ixture\textbf{-o}f\textbf{-}Experts), a progressive training framework that incrementally expands the expert pool during training. Our approach starts from a small (or dense) model and upcycles it into larger MoE models through multiple expansion stages, enabling efficient utilization of both compute and data.
Prior work on routing and load balancing~\citep{fedus2022switch,clark2022unified,zhou2022mixture}, systems-level optimization~\citep{lepikhin2021gshard,gale2023megablocks,rajbhandari2022deepspeed,yan2026scalable,zhang2025comet,jiang2024lancet}, and sparse upcycling~\citep{komatsuzaki2023sparse, nakamura2025drop, he2024upcycling} all fix $E$ throughout training; EMO instead lets $E$ grow with the data, not relying on architectural manipulations beyond standard MoE layers or auxiliary objectives.

However, the effectiveness of EMO hinges on when to expand. 
If training remains at small $E$ for too long, the model may underutilize the data needed by its later expanded capacity; if expansion occurs too early, training incurs the full large-$E$ wall-clock cost for most of the run.
To address this question, we study \textit{optimal stage-wise token allocation} through fitting a sparsity scaling law that explicitly model the effect of expert count $E$. Concretely,  we run a series of scaling-law experiments, calibrated on a sweep $E \in \{2, \ldots, 256\}$ on our data and architecture. Under this, we predict the performance of each stage and assign precisely the data every expert capacity justifies. The resulting schedule front-loads cheap small-$E$ training and reserves large-$E$ for the end.

We validate this at scale by progressively expanding a 1.1B dense model into a 9.6B-A1.1B MoE with $E{=}128$ over five stages on 1.92T tokens. EMO reaches a final pretraining loss of 1.017, compared to 0.994 for the fixed-$E{=}128$ baseline (a 2.3\% relative gap) and 1.065 for the fixed-$E{=}64$ baseline, while saving 10\% GPU hours. Downstream evaluation on eight benchmarks spanning reasoning, knowledge, and commonsense shows the same: EMO clearly outperforms the fixed-$E{=}64$ baseline and remains comparable to the fixed-$E{=}128$ ceiling. 


\section{Preliminary}
\subsection{Mixture of Experts}
MoE models extend standard transformer architectures by replacing the feed-forward network (FFN) with $E$ experts and a routing function that sparsely selects a subset for each token~\citep{shazeer2017outrageously, fedus2022switch}. Given token representation $x$, the router assigns probabilities over experts and activates only the top-$k$:

\begin{equation*}
\text{MoE}(x) = \sum_{i \in \mathcal{K}(x)} p_i(x) \cdot E_i(x),
\end{equation*}

where $\mathcal{K}(x) \subset \{1, \ldots, E\}$ with $|\mathcal{K}(x)| = k$ denotes the selected expert indices and $p_i(x)$ are the normalized routing weights. Since $k \ll E$ in practice, per-token floating-point operations (FLOPs) scale with $k$, enabling parameter count to grow quasi-independently of computation.

\begin{figure}[t]
    \centering
    \begin{minipage}[t]{0.50\textwidth}
        \centering
        \includegraphics[width=\linewidth]{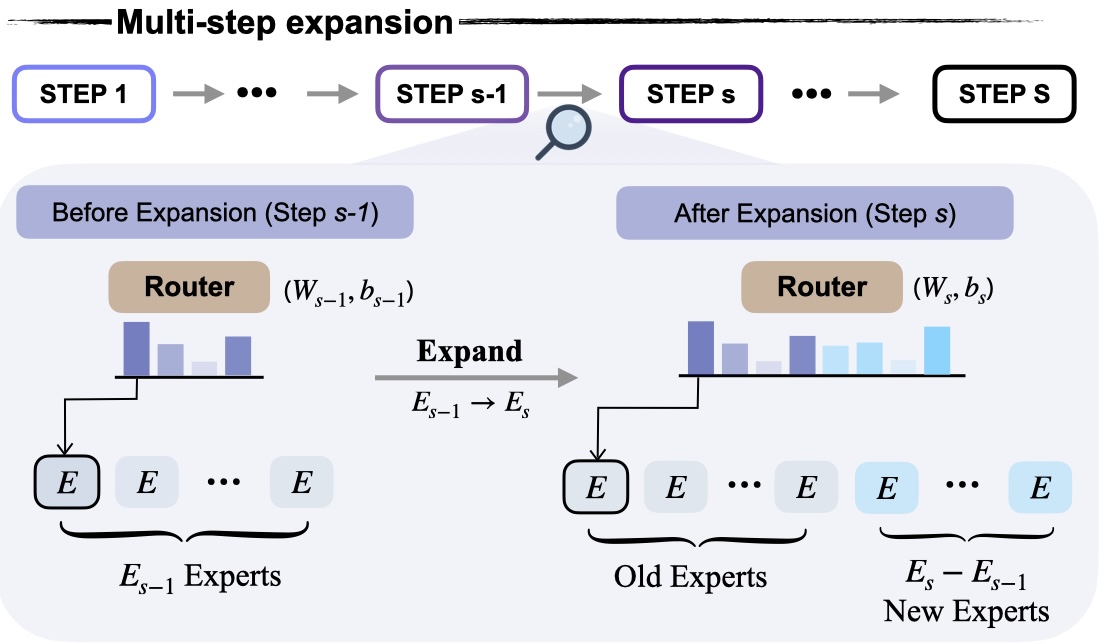}
        \caption{Overview of EMO. EMO performs multi-step expansions; each step we increase model's total expert number with appropriate initialization for new experts and routers.}
        \label{fig:expand_visual}
    \end{minipage}\hfill
    \begin{minipage}[t]{0.48\textwidth}
        \centering
        \includegraphics[width=\linewidth]{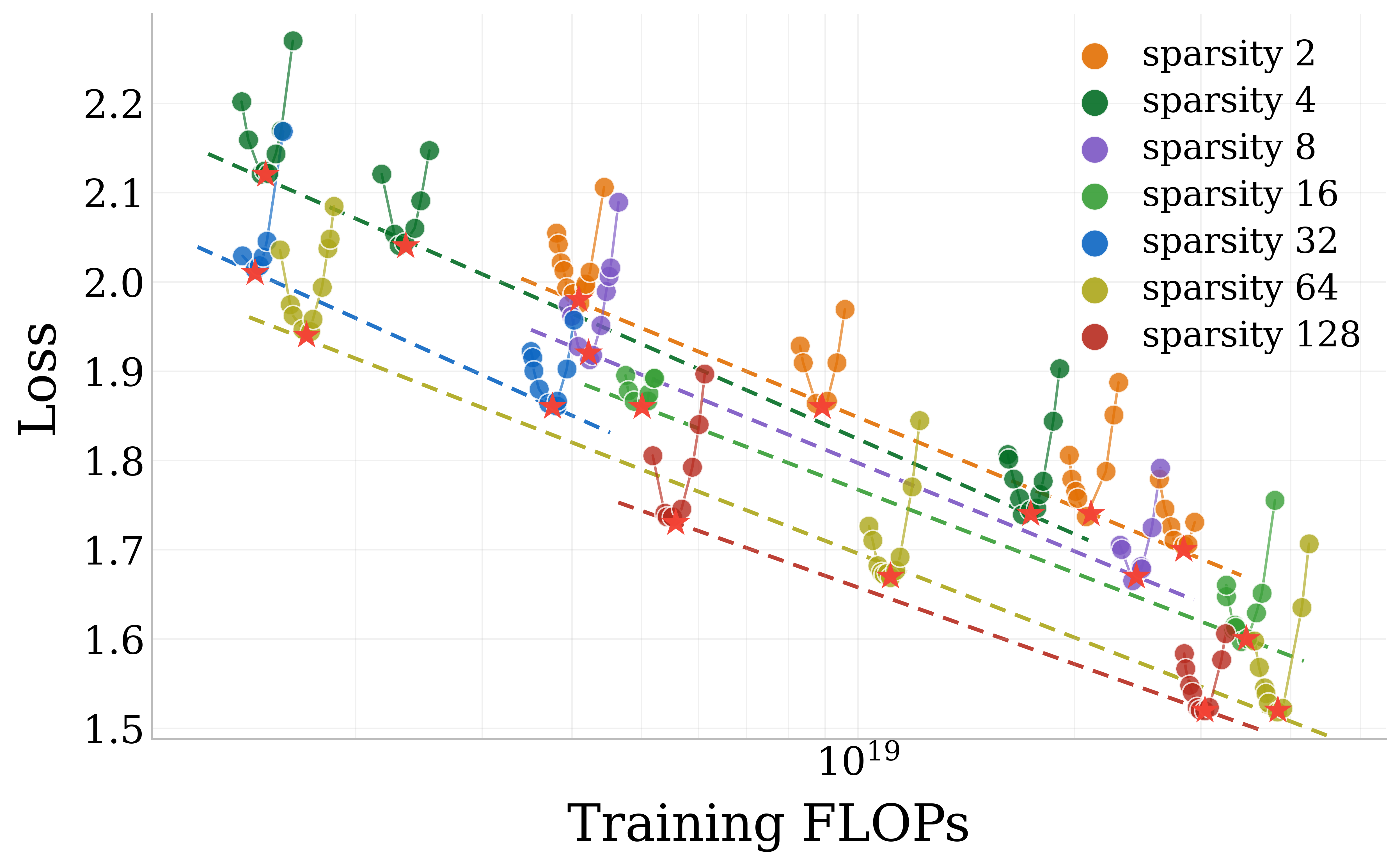}
        \captionof{figure}{Sparsity scaling law. Optimal data–compute changes with E, motivating a sparsity-aware token allocation. We leverage this to derive per-stage optimal token allocations.}
        \label{fig:sparsity}
    \end{minipage}
    \vspace{-3mm}
\end{figure}

\subsection{The MoE Efficiency Paradox}\label{paradox}
Unlike dense transformers, where every parameter participates in every forward pass and FLOPs tracks training wall-clock closely, MoE breaks this correspondence. Per-step FLOPs follow:
\begin{equation*}
    \text{FLOPs}_{\text{step}} \approx 6 \cdot B \cdot N_{\text{act}},
\end{equation*}
where $B$ is batch size and the activated parameter count $N_{\text{act}}$ scales with the number of active experts $k$ rather than the total expert pool $E$. Wall-clock time, however, grows with $E$ through three system-level overheads that scale with the expert pool: all-to-all communication from expert parallelism~\citep{lepikhin2021gshard, rajbhandari2022deepspeed}, aggregate memory for parameters, gradients, and optimizer states~\citep{fedus2022switch}, and GPU underutilization from small 
per-expert GEMMs and routing kernels~\citep{gale2023megablocks, yan2026scalable}.


Figure~\ref{fig:intro} illustrates this gap: at $N_{\text{act}} = 1.1\text{B}$, increasing $E$ from 8 to 128 inflates wall-clock step time by 1.08$\times$; at $N_{\text{act}} = 4\text{B}$, the same expansion costs 1.72$\times$, and the gap is expected to widen further at larger activated sizes. We refer to this FLOPs--wall-clock mismatch as the \emph{MoE efficiency paradox}.

\subsection{Upcycling} Rather than training MoE from scratch, \emph{sparse upcycling}~\citep{komatsuzaki2023sparse} initializes an MoE from a pre-trained dense checkpoint by replicating the FFN weights across experts:
\begin{equation*}
    E_1(x) := E_2(x)... := E_E (x) := FFN(x)
\end{equation*}
where $E_i$ denotes the $i$-th expert and $E$ is the total number of experts.
This initialization preserves the behavior of the original dense model when experts are identical and routing is uniform.  
However, single-step upcycling can be \emph{aggresive} when experts number is large and is not sufficiently efficient.

\section{EMO: Extendable Mixture of Experts}
To smoothly and efficiently move from small MoE or dense model to large MoE model, EMO introduces a multi-step expansion strategy of expert number $E$ with a fixed activated expert number $k$. The algorithm is described in Algorithm \autoref{alg:emo}. At each stage, the model's expert pool grows from $E_s$ to $E_{s+1}$ experts as in \autoref{fig:expand_visual}, and training continues on additional data with the enlarged capacity.
The framework requires two major design decisions: (1) \emph{when} to expand in each stage, to optimize the balance between performance and costs (\S\ref{sec:scaling_law}); (2) \emph{how} to initialize the expansion (\S\ref{sec:init}). 

\begin{algorithm}[t]
\caption{EMO: Progressive MoE Training}
\label{alg:emo}
\textbf{Input:} Initial model $\theta^{(0)}$ with $E_0$ experts, total token budget $D$, final stage expert $E_S$ \\
\textbf{Input:} Expansion schedule $\{(E_s, d_s)\}_{s=1}^{S}$ \hfill $\triangleright$ Sec.~\ref{sec:scaling_law}: compute optimal token counts \\
\textbf{Output:} Trained model $\theta^{(S)}$ with $E_S$ experts
\begin{algorithmic}[1]
\For{$s = 1$ to $S$}
    \State $\theta^{(s-1)} \leftarrow \textsc{Expand}(\theta^{(s-1)},\; E_{s-1} \to E_s)$ \hfill $\triangleright$ Sec.~\ref{sec:init}
    \State $\theta^{(s)} \leftarrow \textsc{Train}(\theta^{(s-1)},\; d_s \text{ tokens})$ 
\EndFor
\State \Return $\theta^{(S)}$
\end{algorithmic}
\end{algorithm}

\subsection{Expert-Aware Token Allocation}
\label{sec:scaling_law}
Given a total budget of $D$ tokens and expansion schedule $E_1 < E_2 < \cdots < E_S$, how should tokens be distributed across stages? Training too long at small $E_s$ wastes model capacity; expanding too early wastes wall-clock time.  We derive a principled allocation by fitting a scaling law that explicitly models the effect of expert count $E$ on data efficiency, then validate it empirically.

\vspace{-1mm}
\paragraph{Step 1: Fit the scaling law.}
To allocate tokens across expansion stages, we need a loss model that captures how data efficiency changes with expert count $E$. Standard compute-optimal scaling laws~\citep{hoffmann2022training} model loss as $L(N, D)$, treating all models with the same activated size identically -- regardless of the total number of experts $E$:
\begin{equation*}
    L(N, D) = a N^{-\alpha} + b D^{-\beta} + c,
\end{equation*}
where $a, b$ are scale coefficients, $\alpha, \beta$ are scaling exponents, $c$ is the irreducible entropy floor.
This is insufficient for EMO, where the same $N_{\text{act}}$ serves different expert configurations at different stages.
To address this issue, we leverage unified MoE scaling law \cite{ludziejewski2025joint} that jointly models the dependence on activated parameters, total experts, and dataset size:
\begin{equation}
    L(N_{\text{act}}, E, D) = m(E) N_{\text{act}}^{\mu(E)} + n(E) D^{\nu(E)} + c,
\end{equation}
where the coefficients and exponents are explicitly parameterized as functions of $E$:
\begin{equation*}
    m(E) = a E^{\delta}, \quad n(E) = b E^{\omega}, \quad
    \mu(E) = \alpha + \gamma \ln E, \quad
    \nu(E) = \beta + \zeta \ln E
    \label{eq3}
\end{equation*}
The exponent $\nu(E)$ controls how quickly loss decreases with additional tokens, indicating the optimal data-to-parameter ratio shifts as experts are added. 

\begin{figure}[t]
    \centering
    \begin{subfigure}{0.24\linewidth}
        \centering
        \includegraphics[width=\linewidth]{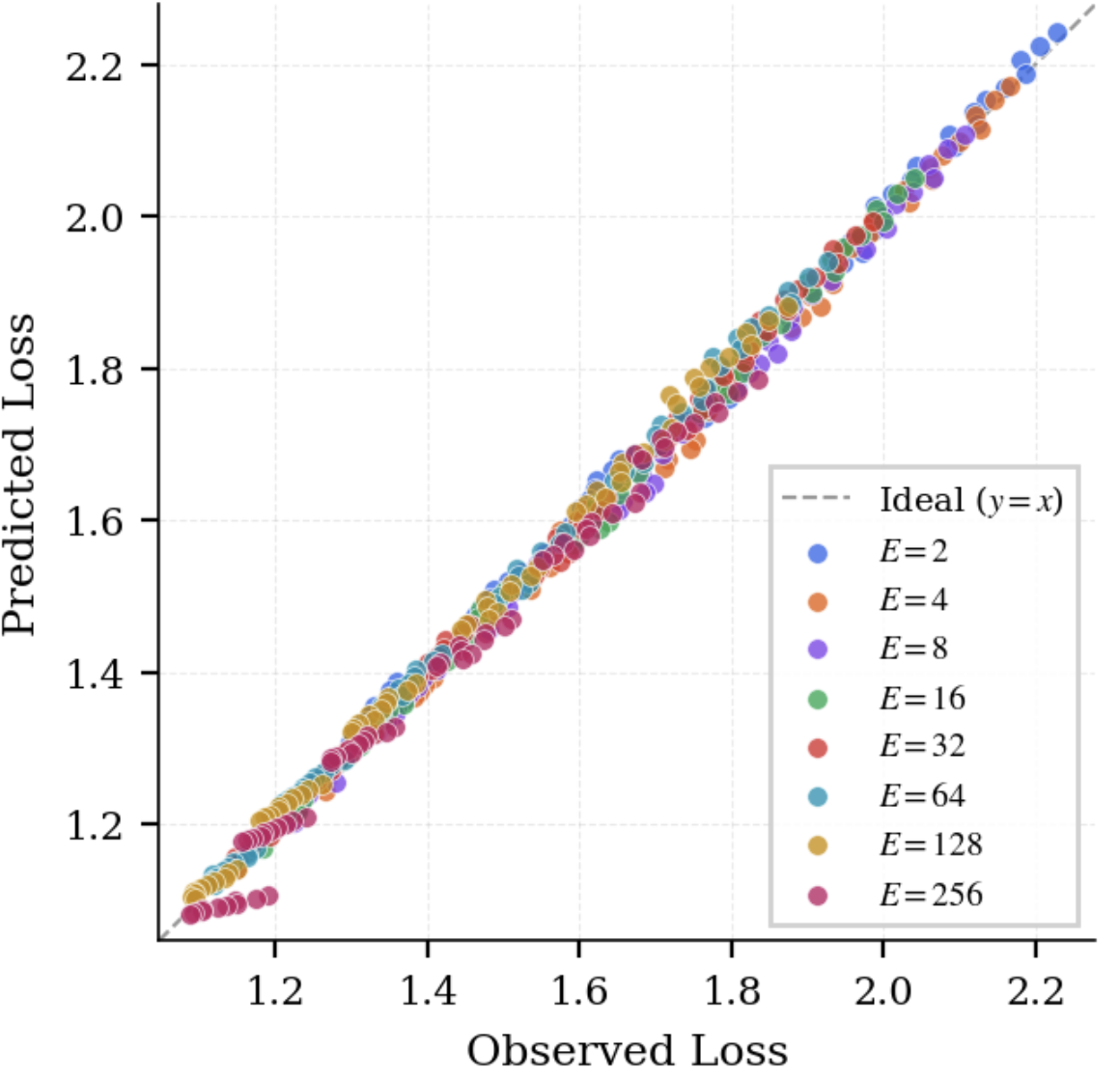}
        \caption{Step 1: Scaling-law fit.}
        \label{fig:scaling_fit}
    \end{subfigure}
    \hfill
    \begin{subfigure}{0.36\linewidth}
        \centering
        \includegraphics[width=\linewidth]{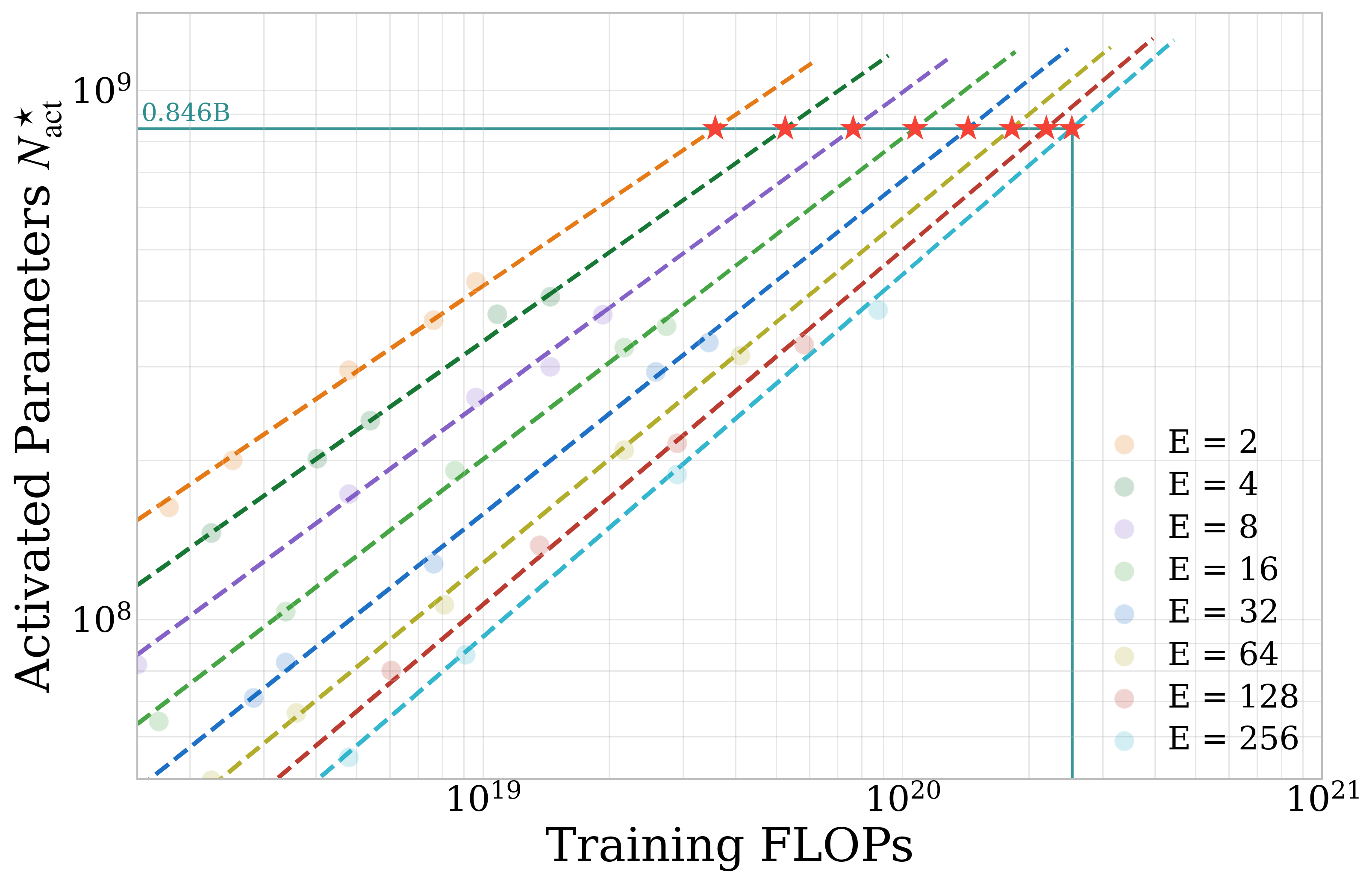}
        \caption{Step 2: Estimate $D^*_s$ across $E$.}
        \label{fig:sparsity_scaling}
    \end{subfigure}
    \hfill
    \begin{subfigure}{0.35\linewidth}
        \centering
        \begin{tcolorbox}[
            colback=blue!4,
            colframe=black,
            boxrule=1.0pt,
            arc=1.5mm,
            left=3pt,
            right=3pt,
            top=5pt,
            bottom=5pt,
            width=\linewidth
        ]
        \centering
        \resizebox{\linewidth}{!}{%
        \begin{tabular}{ccccc}
        \thead{\textbf{Stage}} &
        \thead{\textbf{Sparsity}} &
        \textbf{E} &
        \thead{\textbf{Cum.}\\\textbf{Tokens (B)}} &
        \thead{\textbf{Stage}\\\textbf{Budget}} \\
        \midrule
        1 & 0\%     & 8   & 4.85  & 23.54\% \\
        2 & 50\%    & 16  & 6.80  & 9.47\%  \\
        3 & 75\%    & 32  & 10.20 & 16.50\% \\
        4 & 87.5\%  & 64  & 14.70 & 21.84\% \\
        5 & 93.75\% & 128 & 20.60 & 28.64\% \\
        \end{tabular}
        }
        \end{tcolorbox}
        \caption{Step 3: Compute stage-wise budget $d^*_s$ in our schedule.}
        \label{tab:emo_token_allocation}
    \end{subfigure}
    \caption{Stage-wise, expert-aware token allocation. We study how to optimally allocate tokens in progressive training given fixed activated parameters and token budget. As sparsity-aware scaling law makes progressive training \emph{predictable}, we estimate cumulative per-expert optimal token allocations first, then normalize them into our exapansion schedule with total token budget. }
    \vspace{-3mm}
    \label{fig:loss_prelim}
\end{figure}

With the same data as our main experiments, we fit Eq.~\eqref{eq3} on a grid of small-scale runs varying $E \in \{1, 2, 4, ..., 256\}$, $N_{act}$ and $D$. Figure~\ref{fig:scaling_fit} shows predicted vs.\ observed loss across all configurations, validating the effectiveness of step 1\footnote{The fit achieves $R^2 = 0.9957$ and $RMSE=0.0085$ on the training data and $RMSE=0.0092$ in held-out set.}. Figure~\ref{fig:sparsity} visualizes the sparsity scaling law on our dataset. 

\vspace{-1mm}
\paragraph{Step 2: Compute per-expert optima.} 
As $N_{\text{act}}$ is fixed throughout training, loss depends only on $E_s$ and cumulative tokens $D_s$ at each stage $s$. 
We determine the optimal token allocation across stages using the fitted scaling law 
$L(N_{\text{act}}, E, D)$ and compute-optimal training principles. 
At each stage $s$ with expert count $E_s$, we solve for the \emph{compute-optimal token count} under fixed compute $F$:
\begin{equation}
D_s^* \;=\; \arg\min_D \; L(N_{\text{act}}, E_s, D) 
\quad \text{s.t.} \quad F = 6 N_{\text{act}} D.
\end{equation}
$D_s^{*}$ is the optimal \emph{cumulative} token count if the model were trained entirely with $E_s$ experts. Figure~\ref{fig:sparsity_scaling} presents the resulting $D_s^*$ for each $E$.

\vspace{-1mm}
\paragraph{Step 3: Normalize into a schedule.} We convert cumulative optima into incremental allocations:
\begin{equation}
d_s^* = D_s^* - D_{s-1}^*, \qquad D_0^* = 0.
\label{eq:allocation}
\end{equation}

Given a total token budget $D_{\text{total}}$, we normalize to obtain the final per-stage allocation:
\begin{equation}
d_s = D_{\text{total}} \cdot \frac{d_s^*}{\sum_{s'=1}^{S} d_{s'}^*}.
\end{equation}
Intuitively, $D_s^{*}$ captures the data requirement of a model with $E_s$ experts under optimal scaling, and the differences $d_s^{*}$ reflect how much \emph{additional} data is justified when expanding capacity from $E_{s-1}$ to $E_s$. Normalization ensures the total budget is respected while preserving the relative proportions dictated by the scaling law. Table~\ref{tab:emo_token_allocation} shows the resulting schedule for our main experiment.

\subsection{When to Expand: Validating the Allocation.}
\label{sec:when_to_expand}

The scaling law tells us \emph{how many} tokens each stage deserves, but does the predicted allocation actually sit in a favorable region of the quality--cost landscape? We test this by growing an $E=16$ model to $E=32$ at three fixed fractions of the total token budget (25\%, 50\%, 75\%), bracketing the scaling-law-derived expansion point at approximately 45\%.

\vspace{-1mm}
\begin{wrapfigure}{r}{0.5\textwidth}
\vspace{-15pt}
\includegraphics[width=.5\textwidth]{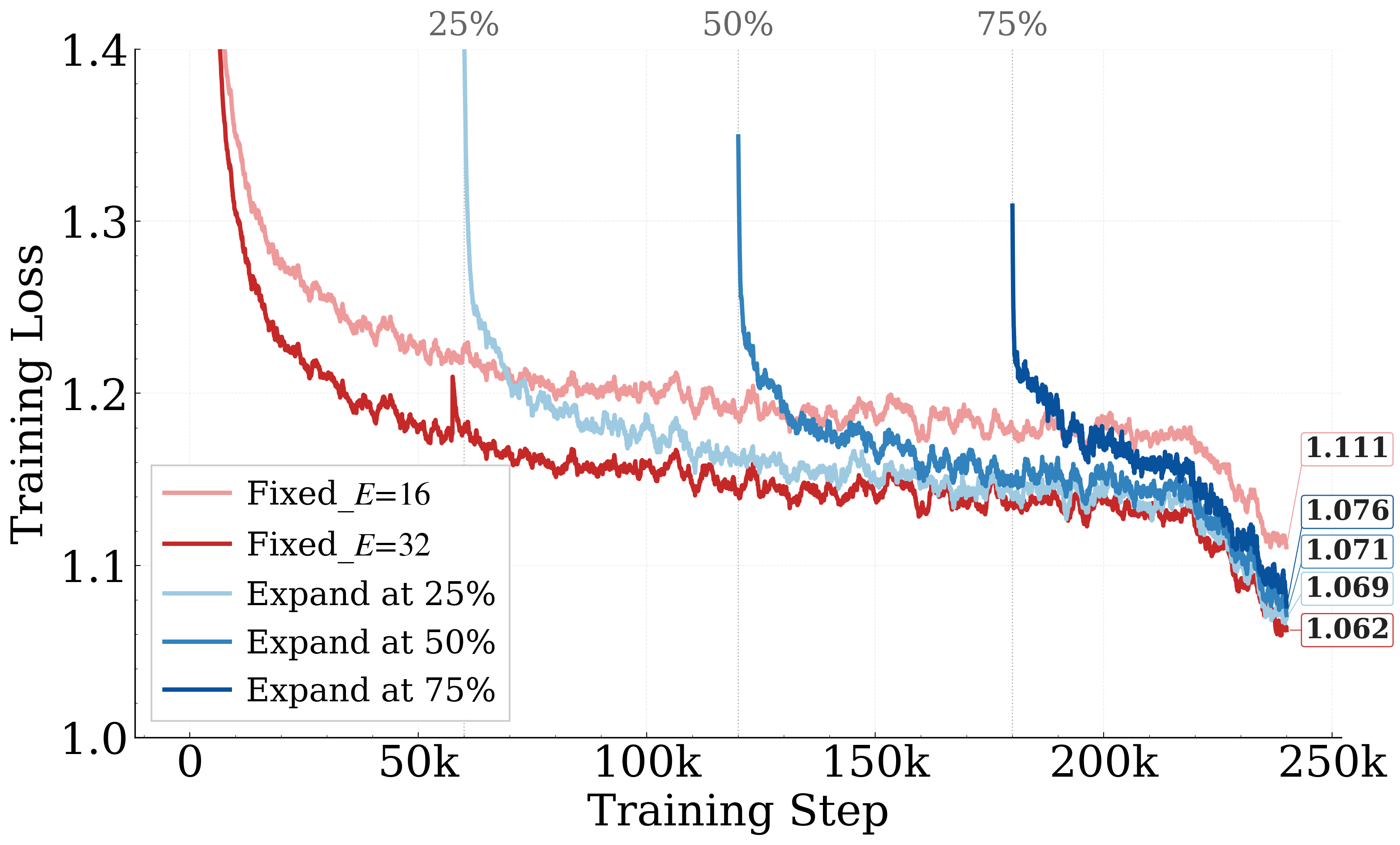}
\caption{Validating Token Allocation: increasing experts $E=16\rightarrow32$ @25\%, 50\% and 70\%. }
\vspace{-2mm}
\label{fig:loss_prelim}
\end{wrapfigure}
\begin{figure}[h]
    \centering
    \includegraphics[width=1\linewidth]{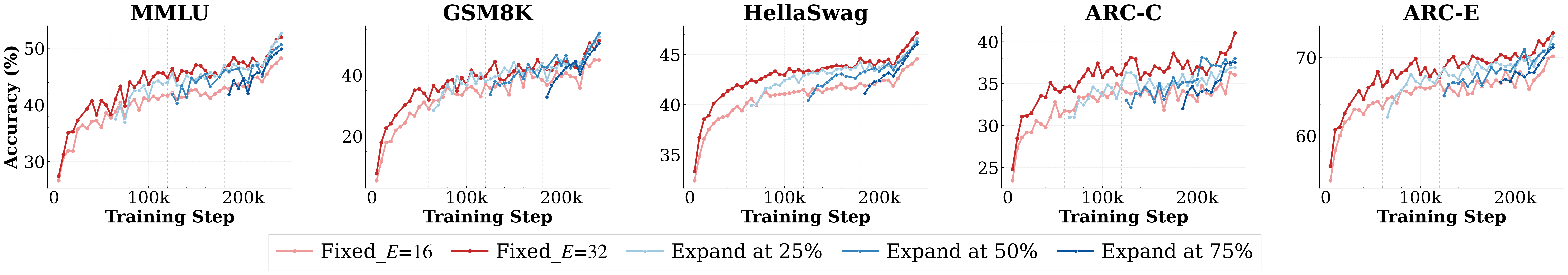}
    \caption{Downstream curves across different expansion timing ($E = 16 \to 32$). Comparing to \textsc{Fixed\_E=32}, EMO@25\% outperforms on both MMLU and GSM8K, performs comparably on HellaSwag and ARC-E. Even EMO@75\% performs much better than \textsc{Fixed\_E=16}. }
    \vspace{-10pt}
    \label{fig:downstream_prelim}
\end{figure}

\paragraph{The scaling law targets the right region.} Figure~\ref{fig:loss_prelim} shows 
the final losses of all three expansions fall between the fixed $E=16$ and fixed $E=32$ baselines. 
Expanding at 25\% achieves the lowest loss (1.069), while expanding at 50\% and 75\% reach 1.071 and 1.076 respectively---each step later in timing costs quality but saves wall-clock time. 
The loss difference between 25\% and 50\% expansion is only 0.002, while the gap widens to 0.007 between 50\% and 75\%. This indicates that the quality--cost curve is relatively flat in the 25--50\% region and steepens beyond 50\%. Our scaling-law-derived allocation at ${\sim}45\%$ falls squarely in this flat region, capturing most of the quality benefit of early expansion while avoiding the full wall-clock cost of the earliest (25\%) schedule. Figure~\ref{fig:downstream_prelim} confirms the same pattern on downstream benchmarks.

\takeaway{
The quality--cost trade-off is favorable near the scaling-law-predicted expansion point: expanding later than $d^*_s$ loses quality rapidly, while expanding much earlier diminishes returns.
}

\subsection{Expert Expansion and Initialization}
\label{sec:init}
Consider an expansion step $s$ that increases the expert number from $E_{s-1}$ to $E_s$, three components must be initialized: the \emph{new expert weights}, the \emph{router weights}, and the \emph{router bias}. Let the model parameters at step $s-1$ be:
\[
\theta_{s-1} = \{\theta_i^{\text{old}}\}_{i=1}^{E_{s-1}}, \quad
W_{s-1} \in \mathbb{R}^{E_{s-1} \times d_r}, \quad
b_{s-1} \in \mathbb{R}^{E_{s-1}}.
\]

After expansion, the parameters become:
\[
\theta_s = \{\theta_i^{\text{old}}\}_{i=1}^{E_{s-1}} \cup \{\theta_j^{\text{new}}\}_{j=E_{s-1}+1}^{E_s},
W_s =
\begin{bmatrix}
W_{s-1} \\
W_{\text{new}}
\end{bmatrix}, \quad
b_s =
\begin{bmatrix}
b_{s-1} \\
b_{\text{new}}
\end{bmatrix},
\]
where $W_{\text{new}} \in \mathbb{R}^{(E_s - E_{s-1}) \times d_r}$ and $b_{\text{new}} \in \mathbb{R}^{E_s - E_{s-1}}$ correspond to the newly added experts.  

Figure~\ref{fig:expand_visual} illustrates how the expansion works. We ablate initialization strategies in \S\ref{sec:ablations} and find that EMO is robust across choices: all configurations converge to similar final loss, with the main difference being the size of the transient spike at expansion. For elegance, in all main experiments, we use Gaussian initialization for new experts and router weights, and reset all router biases to zero. Optimizer states are reset at each expansion with a short learning rate warmup.

\section{Experiments}

\subsection{Experimental Setup}
\vspace{-1mm}
\paragraph{Architecture and training.}
We train decoder-only Transformer MoE models with a fixed
top $k=8$ routing strategy and vary the total expert count $E \in \{8, 16, 32, 64, 128\}$, with activated parameters $1.1B$ with embedding parameters.
\textsc{EMO} proceeds in five stages, each doubling the expert pool: $8 \to 16 \to 32 \to 64 \to 128$. At each expansion boundary, the new stage resumes from the previous stage's last checkpoint and continues training until a stage-specific token budget $D^*_s$. We use a warm-stable-decay (WSD)~\citep{hu2024minicpm} learning rate schedule: all intermediate stages warms-up for 500 steps and train at the constant learning rate, so expansion introduces no discontinuity in the schedule. The learning rate is decayed only in the final stage, beginning at 90\% of total token budget. All models share identical non-expert backbone parameters, optimizer hyperparameters, learning rate schedules, and data mixtures to ensure a controlled comparison. We evaluate EMO under a highly optimized training stack through advanced optimization techniques to ensure the effectiveness of EMO. More details are described in Appendix \ref{app:experimental_setup}.

\vspace{-1mm}
\paragraph{Scaling-law allocation.} Our allocation from Eq.~\eqref{eq:allocation} assigns stage fractions of 23.5\%, 9.5\%, 16.5\%, 21.8\%, and 28.6\% for the $E = 8, 16, 32, 64, 128$ stages (Table~\ref{tab:emo_token_allocation}). This schedule navigates the quality--efficiency trade-off automatically: it front-loads enough tokens at small $E$ to build useful representations, then allocates the majority of the budget to the later, higher-capacity stages where the scaling law predicts the greatest marginal returns.

\vspace{-1mm}
\paragraph{Baselines}
We compare EMO against three from-scratch baselines trained with fixed expert pool $E \in \{16, 32, 128\}$, with the same total token budget and identical hyperparameters, denoted as \textsc{Fixed\_16}, \textsc{Fixed\_32} and \textsc{Fixed\_128}.

\begin{figure}[t]
        \centering
        \includegraphics[width=\linewidth]{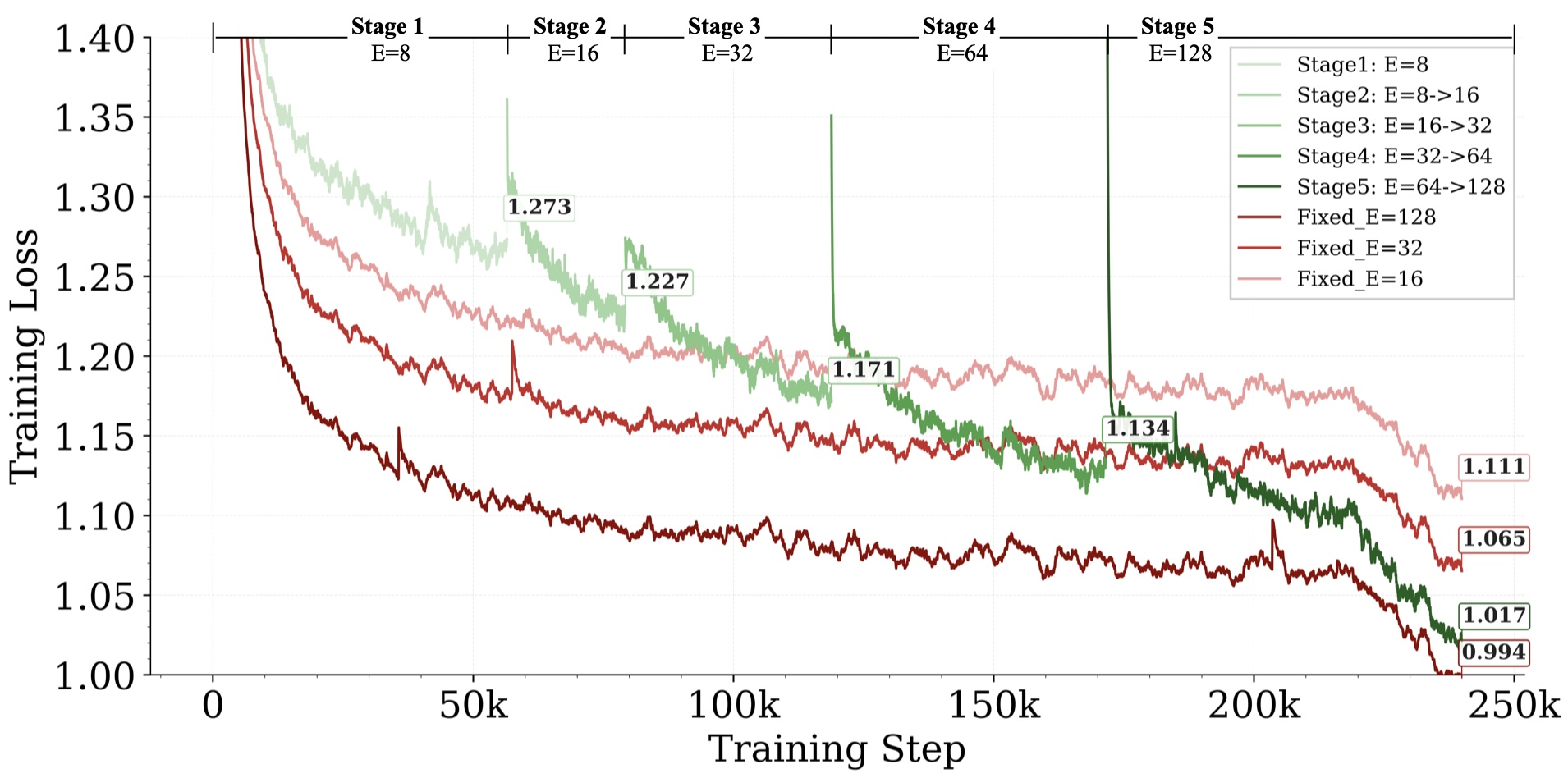}
        \caption{Training-loss comparisons under fixed FLOPs. EMO starts from $E=8$ and progressively expands to $E=128$. EMO reaches a comparable loss as \textsc{Fixed\_E=128} baseline while being more efficient in training and GPU memory. EMO greatly outperforms \textsc{Fixed\_E=32} and \textsc{Fixed\_E=16}.}
        \label{fig:loss_core}
        \vspace{-3mm}
\end{figure}

\begin{figure}[t]
    \centering
    \includegraphics[width=1\linewidth]{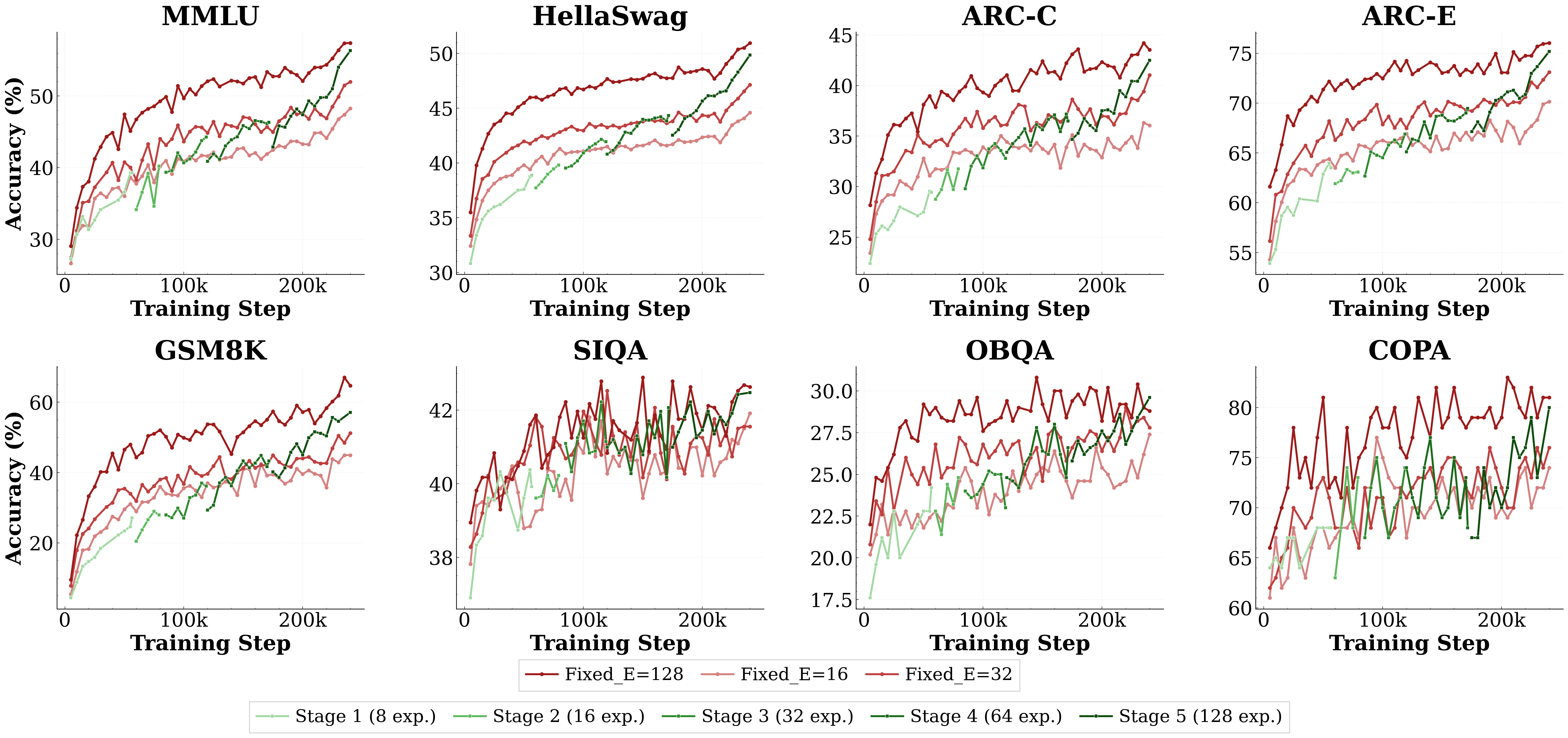}
    \caption{Benchmark curves during training.
We evaluate EMO and fixed-expert baselines on eight downstream benchmarks. EMO is competitive with or stronger than  \textsc{Fixed\_E=128}. Meanwhile, EMO consistently exceeds \textsc{Fixed\_E=32} and \textsc{Fixed\_E=16} in downstream tasks. }
    \label{fig:downstream_core}
    \vspace{-3mm}
\end{figure}

\vspace{-1mm}
\paragraph{Infrastructure.} Training is performed on 32 NVIDIA H200 GPUs with data parallelism. To evaluate EMO under a highly optimized training stack, we use BF16 mixed precision, FlashAttention~3~\citep{shah2024flashattention}, fused transformer-block kernels, and selective activation recomputation for efficiency. We also employ multi-source sequence packing based on a best-fit-decreasing heuristic to minimize padding waste, with document masking. For MoE-specific optimization, we use entropy-based load balancing and recompute router activations to reduce activation memory~\citep{korthikanti2023reducing}.

\vspace{-1mm}
\paragraph{Data.}

\begin{wrapfigure}{r}{0.4\textwidth}
\vspace{-30pt}
\includegraphics[width=\linewidth]{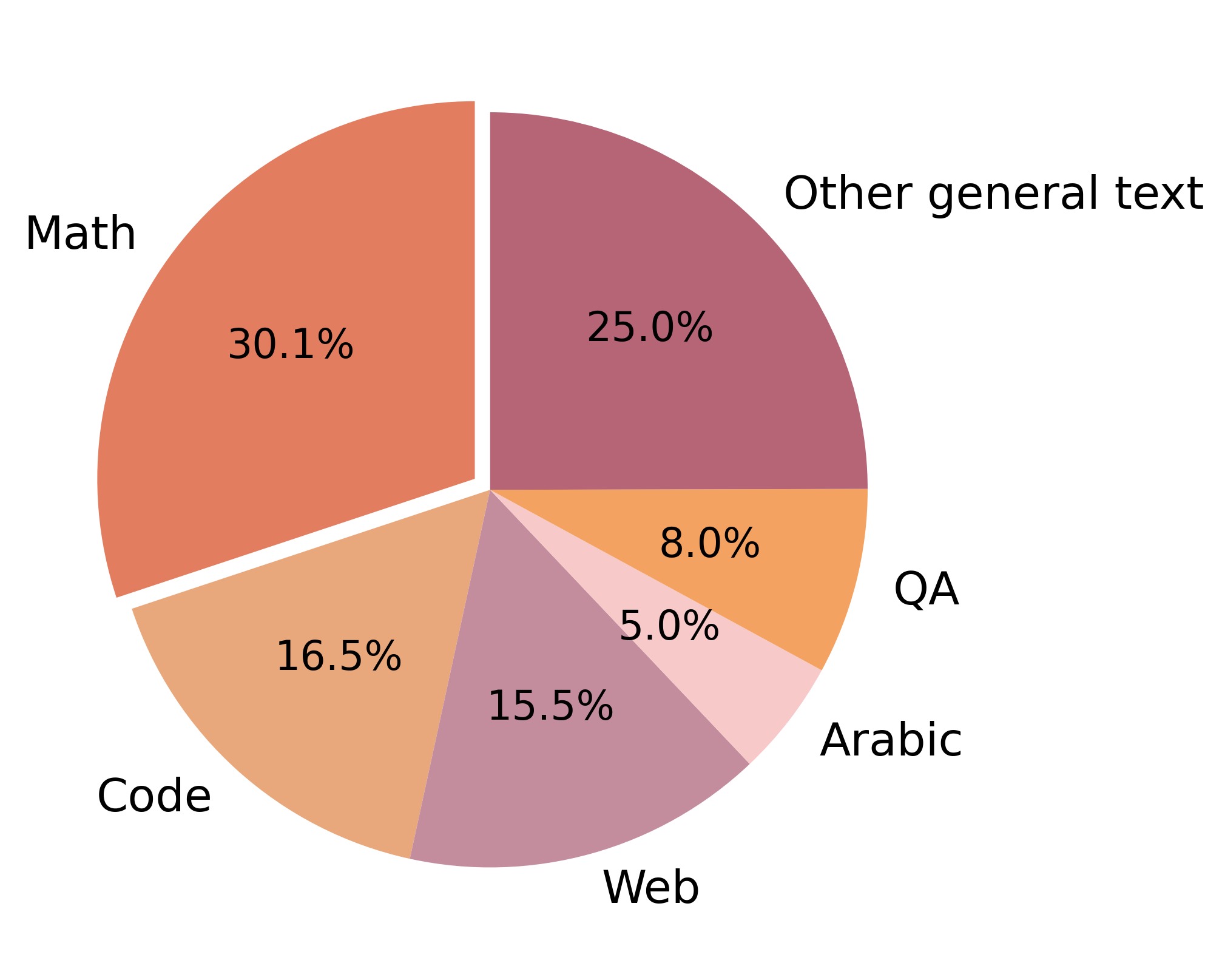}
\vspace{-25pt}
\caption{Training data mix.}
\label{fig:data_mix}
\vspace{-10pt}
\end{wrapfigure}
We pretrain on a mixture of web, code, mathematical, and multilingual corpora following standard large-scale pretraining practices. The total token budget is fixed at $1.92T$ tokens across all runs. 
Validation perplexity is evaluated every 5K steps on held-out web, multilingual, code, academic, and other validation slices.  Downstream evaluation is also run every 5K steps on \textsc{BoolQ}, \textsc{HellaSwag}, \textsc{Natural Questions}, \textsc{PIQA}, \textsc{SIQA}, \textsc{WinoGrande}, \textsc{OpenBookQA}, \textsc{ARC-Easy}, \textsc{ARC-Challenge}, \textsc{RACE}, \textsc{COPA}, \textsc{MMLU}, Arabic MMLU, \textsc{TruthfulQ A}, and \textsc{GSM8K}.

\subsection{Main Results}

Figure~\ref{fig:loss_core} compares EMO with fixed-expert MoE baselines trained under the same total token budget and activated parameter count.
EMO starts from a much smaller expert number ($E=8$ at the beginning), but reaches the same low-loss regime as the \textsc{Fixed\_$E=128$} baseline by the end of training, with a final loss of $1.017$ versus $0.994$.
At the same time, EMO clearly outperforms smaller fixed-expert baselines such as \textsc{Fixed\_$E=16$} and \textsc{Fixed\_$E=32$}, showing that progressive expansion avoids the capacity bottleneck of small expert pools while avoiding the cost of using the largest expert pool throughout training.
Each expansion introduces a transient loss spike, but the loss recovers within roughly $10$K steps, suggesting that the newly added experts are integrated quickly rather than causing persistent optimization instability.
Together, these results show that EMO provides a stable and efficient training trajectory: it uses cheaper small-expert configurations early, then successfully recovers the benefits of large expert capacity in later stages.

We next evaluate whether this upstream training behavior transfers to downstream performance.
Figure~\ref{fig:downstream_core} reports accuracy on multiple benchmarks (See full downstream results and validation perpexity in Appendix~\ref{app:analysis}).
After the final expansion, EMO is competitive with the fixed $E=128$ baseline on most benchmarks and clearly improves over smaller fixed-expert baselines.
On GSM8K, the fixed $E=128$ baseline remains stronger at the final checkpoint, suggesting that some reasoning-heavy benchmarks may benefit more from exposing the large expert pool throughout training.
Overall, these results show that treating MoE capacity as expandable parametric memory improves the quality--cost trade-off of MoE training: EMO retains much of the benefit of large expert pools while substantially reducing the cost of reaching them.

\takeaway{
EMO shows that large-expert MoE performance does not require large-expert MoE cost from the first step
, and progressive expansion recovers most of the final capacity benefit while significantly reducing early-stage training overhead.
}

\vspace{-1mm}
\section{Analysis}
\label{sec:ablations}

\subsection{Expansion Initialization}
We first study how to initialize newly introduced experts and router parameters when expanding the expert pool. \autoref{fig:init_strategy} compares three strategies for the $16 \to 32$ expansion:

\begin{figure}[t]
    \centering
    \begin{minipage}{0.48\linewidth}
        \centering
        \includegraphics[width=\linewidth]{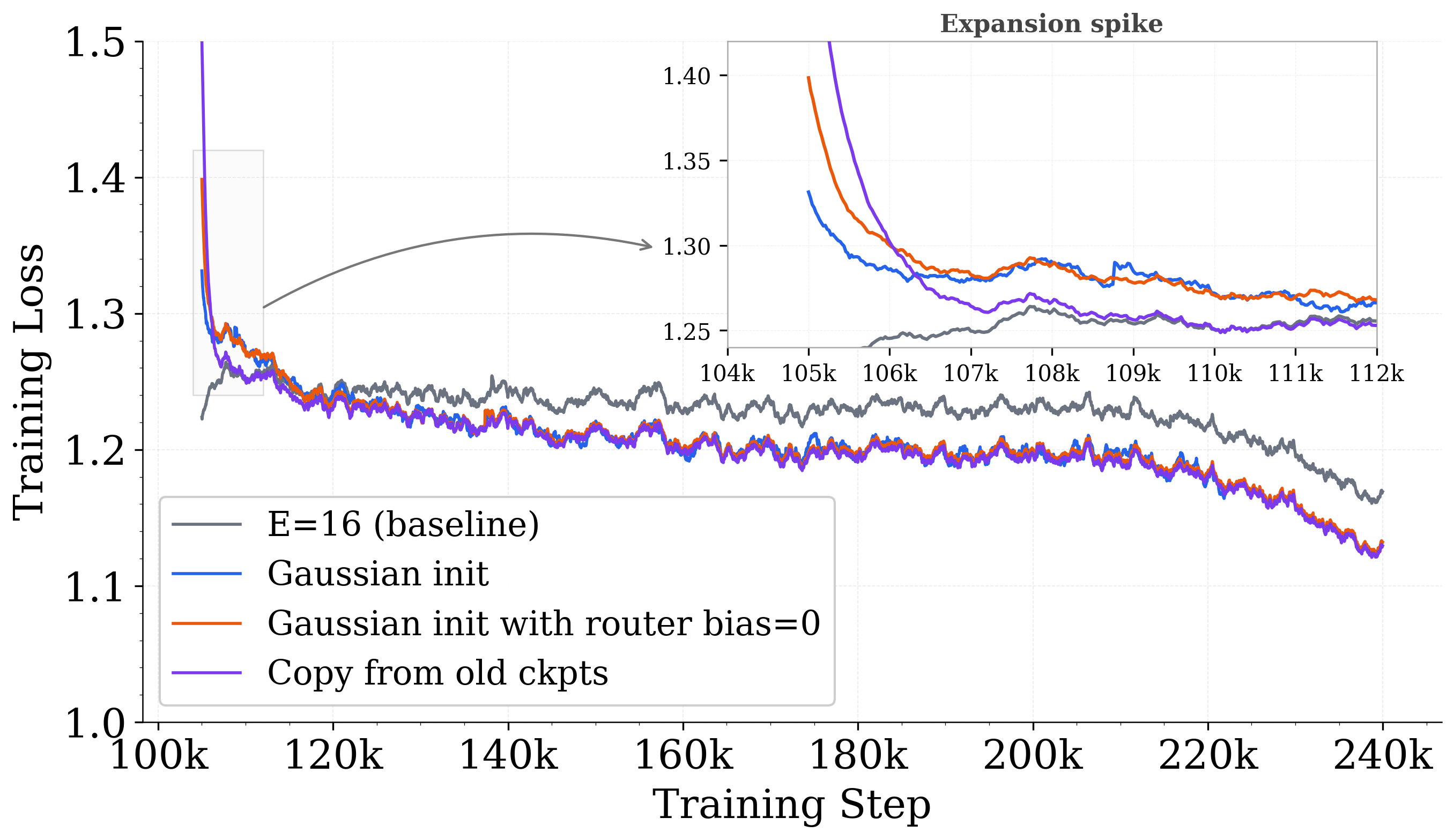}
        \caption{Expert \& Router Initialization Strategy. EMO is robust to initialization strategy. All configurations converge to similar final loss; the choice mainly affects the transient spike.}
        \label{fig:init_strategy}
        \vspace{-10pt}
    \end{minipage}
    \hfill
    \begin{minipage}{0.48\linewidth}
        \centering
        \includegraphics[width=\linewidth]{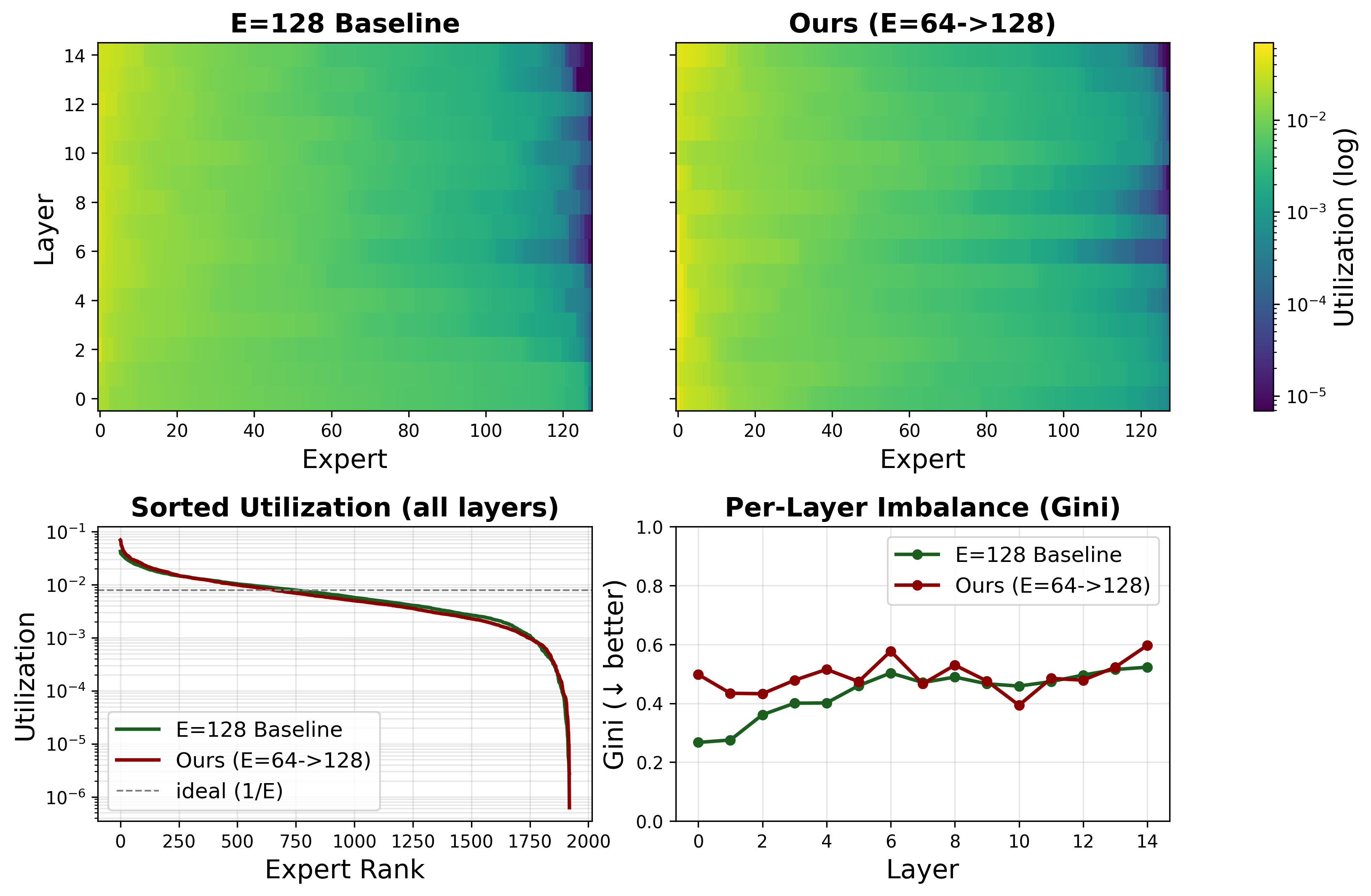}
        \caption{Expert utilization on validation data. Top: per-layer × per-expert utilization; bottom-left: utilization curves aggregate all layers; bottom-right: per-layer Gini summarizes imbalance (0 = uniform, 1 = collapsed).}
        \label{fig:expert_util}
    \end{minipage}
    \vspace{-10pt}
\end{figure}

\begin{itemize}[itemsep=2pt, parsep=0pt, topsep=0pt, partopsep=0pt, leftmargin=*]
    \item \textit{Gaussian init}: randomly initialize the new experts and router parameters.
    \item \textit{Gaussian init with router bias=0}: randomly initialize new experts and router weights, while resetting router biases to zero to clear prior load-balancing state.
    \item \textit{Copy from old ckpts}: initialize new experts and router parameters from existing checkpoints. 
\end{itemize}
For reference, the gray curve shows the FIXED\_E$=16$ baseline without expansion. All three expansion strategies substantially outperform the non-expanded $E=16$ baseline, indicating that the benefit of EMO is not tied to a single fragile initialization choice. The strategies differ only marginally in final training loss given a sufficient token budget. At the expansion boundary, all methods exhibit a transient loss spike, but the spike disappears quickly during warmup and training remains stable, suggesting that new experts and routers learn fast even when starting from scratch.

In particular, \emph{Copy from old ckpts} starts with the highest spike but stabilizes fastest. The bias-reset variant produces a larger spike than plain Gaussian init because resetting the router bias discards the load-balancing behavior learned in the previous stage and abruptly changes token assignments; both nonetheless converge to similar loss after warmup. As our token budget far exceeds the spike regime, we use the bias-reset variant in our main experiments for simplicity. 

When expanding the expert pool, optimizer states from the previous stage can either be carried over together with the model parameters or reset for the new stage. We find the improvement from carrying them over is marginal: differences vanish after roughly 500 warmup steps. We therefore reset optimizer states to avoid dimension-mismatch handling when new experts add rows to Adam moment buffers across all expansion stages.

\takeaway{
    EMO is robust to both initialization and optimizer-state handling at expansion boundaries: expert learning is fast, and the choice mainly affects the size of the transient spike.
}

\subsection{Expert utilization comparison} 
We measure routing balance on 5K validation sequences by collecting tokens per expert from every MoE router. 
Both \textsc{Fixed\_$E=128$} and our progressively expanded model
exhibit similar distributions (\autoref{fig:expert_util} top): experts in middle and last few layers receive relatively larger load. Quantitatively, the Gini coefficient is 0.44 for the baseline and 0.50 for our expanded model, a $\sim$14\% relative gap that is concentrated in the middle layers (\autoref{fig:expert_util} bottom right), where newly initialized experts inherit slightly more uneven routing weights from the parent expert during expansion. 
Crucially, no layer of the expanded model exhibits collapse (Gini $<$ 1). The expert utilization curves (\autoref{fig:expert_util} bottom left) overlap closely.Progressive expansion thus achieves similar routing behavior to the baseline, despite the new experts being trained for shorter time.

\subsection{MoE as expandable memory}
\begin{figure}[t]
    \centering
    \includegraphics[width=1\linewidth]{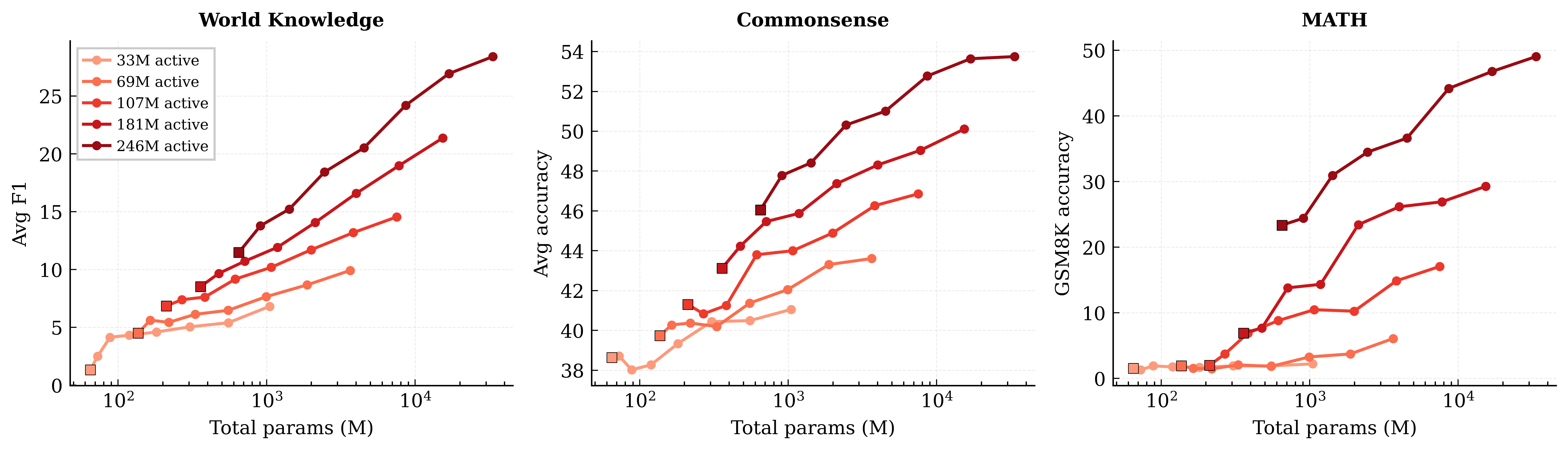}
    \caption{MoE as expandable memory. We evaluate parts of our scaling law MoE models on several world knowledge benchmarks (e.g.,TriviaQA~\citep{joshi2017triviaqa}, NQ~\citep{kwiatkowski2019naturalquestions}).We evaluate multiple commonsense benchmarks including HellaSwag~\citep{zellers2019hellaswag}, WinoGrande~\cite{sakaguchi2021winogrande} etc.; math is evalated on GSM-8K~\cite{cobbe2021gsm8k}. }
    \label{fig:moe_as_memory}
    \vspace{-3mm}
\end{figure}

EMO is motivated by viewing MoE as expandable memory: increasing the expert pool enlarges the number of addressable parameters, improving the model's ability to store and retrieve knowledge. 
We test this in Figure~\ref{fig:moe_as_memory} by varying $E$ from 2 to 256 under different activated-parameter budgets. On memory-intensive tasks such as TriviaQA, larger expert pools consistently improve performance, supporting the role of experts as additional parametric memory. Similar trends appear on commonsense reasoning tasks, while GSM8K benefits more clearly only when total parameter count is sufficiently large. This also explains why the fixed $E=128$ baseline remains stronger on GSM8K: exposing the full expert pool throughout training may give reasoning skills more time to organize across experts. These results support as strong foundations and motivations for EMO: if experts function as addressable parametric memory, then it is not necessary to expose the full memory capacity from the beginning of training. Instead, EMO gradually expands this memory as training progresses, reducing early-stage overhead while preserving the benefits of large expert pools.

\section{Related Work}
\label{sec:related_work}

\paragraph{Sparse MoE models and routing.}
Mixture-of-Experts models increase parameter capacity by routing each token to a sparse subset of experts~\citep{shazeer2017outrageously, lepikhin2021gshard, fedus2022switch}.
Large-scale MoE language models such as GLaM, Mixtral, DeepSeekMoE, DeepSeek-V2, and OLMoE demonstrate that sparse activation can scale model capacity efficiently~\citep{du2022glam, jiang2024mixtral, dai2024deepseekmoe, liu2024deepseek, muennighoff2025olmoe}.
A large body of work improves routing behavior and expert utilization, including load-balancing losses, expert-choice routing, stable transfer designs, and auxiliary-loss-free balancing~\citep{fedus2022switch, zhou2022mixture, zoph2022st, wang2024auxiliary, shi2024unchosen}.
These methods improve how tokens are assigned to a fixed expert pool.
EMO instead addresses when the expert pool itself should grow during training.

\paragraph{MoE upcycling.}
Sparse upcycling converts a pretrained dense model into an MoE by replacing dense FFN layers with expert layers and copying dense weights into multiple experts~\citep{komatsuzaki2023sparse}, reducing the cost of training sparse models from scratch. Recent work improves dense-to-MoE conversion through expert re-initialization~\citep{nakamura2025drop}, virtual-group initialization and weight scaling for fine-grained MoEs~\citep{he2024upcycling}, domain-specialized branches~\citep{sukhbaatar2024branchtrain}, and instruction-tuning upcycling~\citep{hui2025upit}. These methods primarily focus on constructing a fixed MoE, whereas EMO is complementary: it repeatedly expands the expert pool during pretraining and allocates data across expansion stages.

\paragraph{Systems optimizations for MoE training.}

Expert-parallel MoE training requires all-to-all communication to dispatch and combine tokens across devices~\citep{lepikhin2021gshard, rajbhandari2022deepspeed}. Systems work reduces this overhead through optimized dispatchers, communication--computation overlap, grouped GEMM, block-sparse computation, kernel fusion, and memory optimizations~\citep{gale2023megablocks, zhao2025deepep, deepseekv3, yan2026scalable, korthikanti2023reducing}. Early systems such as FastMoE and Tutel made distributed MoE training practical~\citep{fastmoe2021,tutel2022}, while MegaBlocks and Megatron-Core MoE further improve sparse computation and routing efficiency~\citep{gale2023megablocks, yan2026scalable}. These methods lower the cost of a fixed MoE configuration, whereas EMO is orthogonal: it reduces training cost by delaying expensive large-expert configurations until later stages.

\section{Conclusion}

We introduced EMO, a simple progressive training framework for Extendable Mixture-of-Experts models. Motivated by the MoE efficiency paradox, EMO avoids committing to a large expert pool from the start and instead grows expert capacity during training. With a sparsity-aware token allocation strategy, EMO achieves a favorable quality--cost trade-off, approaching fixed large-expert baselines while using cheaper intermediate configurations. It is also robust to expansion initialization choices and maintains broad expert utilization after expansion.
Our study has two main limitations. First, the scaling-law fit does not explicitly model optimization hyperparameters such as learning rate, batch size, or optimizer settings. Second, our experiments remain smaller than frontier-scale MoE systems. Scaling EMO to larger activated and total parameter regimes is an important direction for future work.

\newpage
\bibliographystyle{plainnat}
\bibliography{references}

\newpage
\beginappendix
\appendix

\section{Preliminaries}
\subsection{Notation}

\noindent To aid readability, we provide a list of key symbols used throughout this paper.

\begin{table}[ht]
\centering
\begin{tabular}{@{}ll@{}} 
\toprule
\textbf{Symbol} & \textbf{Description} \\ \midrule
$N$               & Total number of model parameters \\
$N_{act}$             & Active number of model parameters \\
$L$               & Pretraining Loss \\
$F$               & Training compute budget (in FLOPs) \\
$E$               & Expansion factor (number of experts per MoE layer) \\
$K$               & Number of selected experts per token  \\
$D$               & Dataset size (number of training tokens)  \\
$N^*_{act}$       & Compute-optimal active number of parameters \\
$D^*$             & Compute-optimal training token size \\
$E_s$               & total number of experts at expansion stage $s$ \\
$d^*_s$             & incremental training token size at expansion stage $s$ \\
$S$               & Number of total expansion stages \\
$\alpha, \beta, \gamma, \zeta, \delta, a, b, c, m, n$ & Coefficients in the parametric scaling law equation \\ \bottomrule
\end{tabular}
\label{tab:symbols} 
\end{table}

\section{Detailed Experiment Setup}
\label{app:experimental_setup}
\subsection{Main experiment setup}

All main experiments use decoder-only Transformer MoE models with identical dense backbone architecture and training hyperparameters. The model has 16 Transformer layers, hidden dimension 2048, 16 query heads, 4 key-value heads, SwiGLU feed-forward blocks, RMSNorm, RoPE with base 100000, and one dense layer before the MoE layers. Each MoE layer uses one shared expert and a routed expert pool with top-$k=8$ activated routed experts per token. The routed expert hidden dimension is 768, the router uses a sigmoid score function with router bias enabled, and entropy-based load balancing is applied with coefficient $10^{-4}$. The global batch size is 8M tokens, the context length is 8192, and the total training budget is 240K optimizer steps, corresponding to 1.92T tokens. We use AdamW with $\beta_1=0.9$, $\beta_2=0.95$, $\epsilon=10^{-8}$, weight decay 0.05, gradient clipping at 1.0, and peak learning rate $9\times10^{-4}$. The learning-rate schedule is warm-stable-decay: 2K warmup steps, a stable phase, and a linear decay beginning at 90\% of the total token budget to a final learning-rate ratio of 0.01. At every expansion boundary, the model resumes from the previous stage checkpoint, newly introduced experts and router rows are Gaussian initialized, router biases are reset to zero, optimizer states are reset, and a 500-step expansion warmup is applied. All downstream evaluations are conducted using the LM Evaluation Harness~\cite{eval-harness} under its default task templates. Since all evaluated models are base models without instruction tuning, we use greedy decoding (temperature $=0.0$) across all evaluation tasks~\cite{shi2024thorough}.

\label{app:fit}
\subsection{Scaling law experiment setup}
The scaling-law experiments use the same tokenizer, data mixture, context length, optimizer family, router design, and validation protocol as the main runs, but sweep smaller activated model sizes and expert counts to make the fit computationally tractable.  We train a grid of MoE models with $E\in\{2,4,8,16,32,64,128,256\}$. For these scaling runs, top-$k$ is fixed at 2, so changing $E$ changes the total expert pool while keeping sparse activation controlled.

\subsection{Scaling law fitting}

We fit Eq.~\eqref{eq3} using the collected validation losses, varying $N_{\mathrm{act}}$, $E$, and $D$.  Following \citet{hoffmann2022training}, the coefficients are optimized with LBFGS under a Huber loss with threshold 0.01, and the final model selection is performed by grid search over initialization and coefficient constraints.  The resulting fit achieves $R^2=0.9957$ and RMSE 0.0085 on the training points, with held-out RMSE 0.0092.


\section{Additional Analysis}
\label{app:analysis}
\subsection{Downstream Evaluations}
Tables~\ref{tab:final_ckpt} and~\ref{tab:core_downstream_all} complement the validation-loss curves with downstream evaluation.  The preliminary table compares different $E=16\rightarrow32$ expansion timings, while the core table compares the full multi-stage EMO run against fixed-expert baselines.  Across tasks, the downstream results follow the same pattern as the loss curves: progressive expansion is consistently stronger than fixed small-expert baselines and is competitive with the fixed large-expert baseline, especially when accounting for the wall-clock savings from not training with the largest expert pool from the beginning.

\begin{table*}[h]
\centering
\scriptsize
\setlength{\tabcolsep}{3.5pt}
\resizebox{\textwidth}{!}{%
\begin{tabular}{lccccccccccccccc}
\toprule
\textbf{Model} & \textbf{MMLU} & \textbf{HellaSwag} & \textbf{ARC-C} & \textbf{ARC-E} & \textbf{PIQA} & \textbf{BoolQ} & \textbf{GSM8K} & \textbf{SIQA} & \textbf{OBQA} & \textbf{COPA} & \textbf{WinoGrande} & \textbf{NQ (EM)} & \textbf{TriviaQA (EM)} & \textbf{RACE-H} & \textbf{RACE-M} \\
\midrule
FIXED\_$E=16$ & 48.27 & 44.59 & 36.05 & 70.15 & 73.01 & 59.02 & 44.96 & 41.91 & 27.40 & 74.00 & 58.48 & 9.11 & 22.58 & 40.05 & \textbf{57.17} \\
FIXED\_$E=32$ & \textbf{51.98} & \textbf{47.14} & \textbf{41.03} & \textbf{73.11} & 73.78 & \textbf{71.71} & 51.25 & 41.56 & 27.80 & \textbf{76.00} & 60.46 & \textbf{11.86} & \textbf{27.52} & 39.59 & 55.15 \\
Expand @25\% & \textbf{52.70} & 46.63 & 36.91 & 72.69 & \textbf{73.83} & 70.09 & 52.62 & 41.76 & \textbf{29.00} & 75.00 & 60.93 & 10.78 & 27.03 & \textbf{40.57} & 55.50 \\
Expand @50\% & 50.66 & 46.27 & 38.03 & 71.67 & 73.78 & 69.42 & \textbf{53.75} & 41.66 & 27.80 & 75.00 & 60.85 & 10.53 & 26.49 & 40.25 & 55.22 \\
Expand @75\% & 49.88 & 46.01 & 37.51 & 71.25 & \textbf{73.83} & 66.33 & 50.34 & \textbf{42.17} & 26.60 & 73.00 & \textbf{61.64} & 10.06 & 24.93 & 39.85 & 54.60 \\
\bottomrule
\end{tabular}%
}
\caption{Downstream performance of expansion timing experiments, evaluated at the final checkpoint (240k steps).}
\label{tab:final_ckpt}
\end{table*}
\begin{table*}[h]
\centering
\small
\setlength{\tabcolsep}{3.5pt}
\resizebox{\textwidth}{!}{%
\begin{tabular}{lccccccccccccccc}
\toprule
\textbf{Model} 
& \textbf{MMLU} 
& \textbf{HellaSwag} 
& \textbf{ARC-C} 
& \textbf{ARC-E} 
& \textbf{PIQA} 
& \textbf{BoolQ} 
& \textbf{GSM8K} 
& \textbf{SIQA} 
& \textbf{OBQA} 
& \textbf{COPA} 
& \textbf{WinoGrande} 
& \textbf{NQ (EM)}
& \textbf{TriviaQA (EM)}
& \textbf{RACE-H} 
& \textbf{RACE-M} \\
\midrule
FIXED\_$E$=16 
& 48.27 & 44.59 & 36.05 & 70.15 & 73.01 & 59.02 & 44.96 & 41.91 & 27.40 & 74.00 & 58.48 & 9.11 & 22.58 & 40.05 & \textbf{57.17} \\
FIXED\_$E$=32 
& 51.98 & 47.14 & 41.03 & 73.11 & 73.78 & 71.71 & 51.25 & 41.56 & 27.80 & 76.00 & 60.46 & 11.86 & 27.52 & 39.59 & 55.15 \\
FIXED\_$E$=128 
& \textbf{57.40} & \textbf{50.95} & \textbf{43.53} & \textbf{76.03} & \textbf{76.44} & \textbf{74.89} & \textbf{64.73} & \textbf{42.63} & 28.80 & \textbf{81.00} & \textbf{63.06} & \textbf{16.34} & \textbf{39.11} & 40.31 & 56.20 \\
\midrule
Stage 1 ($E=8$) 
& 39.29 & 38.88 & 29.44 & 63.51 & 69.91 & 66.85 & 27.14 & 39.92 & 24.20 & 68.00 & 56.99 & 5.65 & 14.62 & 37.36 & 52.51 \\
Stage 2 ($8{\to}16$) 
& 40.21 & 39.81 & 31.76 & 63.09 & 70.13 & 67.80 & 27.98 & 40.23 & 24.80 & 73.00 & 56.20 & 5.24 & 13.98 & 37.91 & 51.95 \\
Stage 3 ($16{\to}32$) 
& 44.27 & 41.93 & 32.79 & 66.89 & 71.27 & 69.72 & 36.16 & 41.15 & 23.00 & 74.00 & 57.85 & 7.23 & 17.76 & 38.51 & 53.20 \\
Stage 4 ($32{\to}64$) 
& 46.34 & 44.33 & 36.48 & 69.47 & 73.50 & 70.52 & 43.29 & 42.07 & 26.60 & 68.00 & 58.56 & 8.64 & 22.21 & 39.71 & 54.81 \\
Stage5 ($64{\to}128$) 
& 56.34 & 49.86 & 42.49 & 75.18 & 74.81 & 70.40 & 57.09 & 42.48 & \textbf{29.60} & 80.00 & 61.88 & 13.35 & 33.77 & \textbf{40.39} & 55.92 \\
\bottomrule
\end{tabular}%
}
\caption{
Main downstream evaluation at the last checkpoint. All numbers are accuracy (\%). Across tasks, the downstream results follow the same pattern as the loss curves: progressive expansion is consistently stronger than fixed small-expert baselines and is competitive
with the fixed large-expert baseline, especially when accounting for the wall-clock savings from not training with the largest expert pool from the beginning.}
\label{tab:core_downstream_all}
\end{table*}

\subsection{Training Loss}
\begin{figure}[h]
    \centering
    \begin{minipage}{0.48\linewidth}
        \centering
        \includegraphics[width=\linewidth]{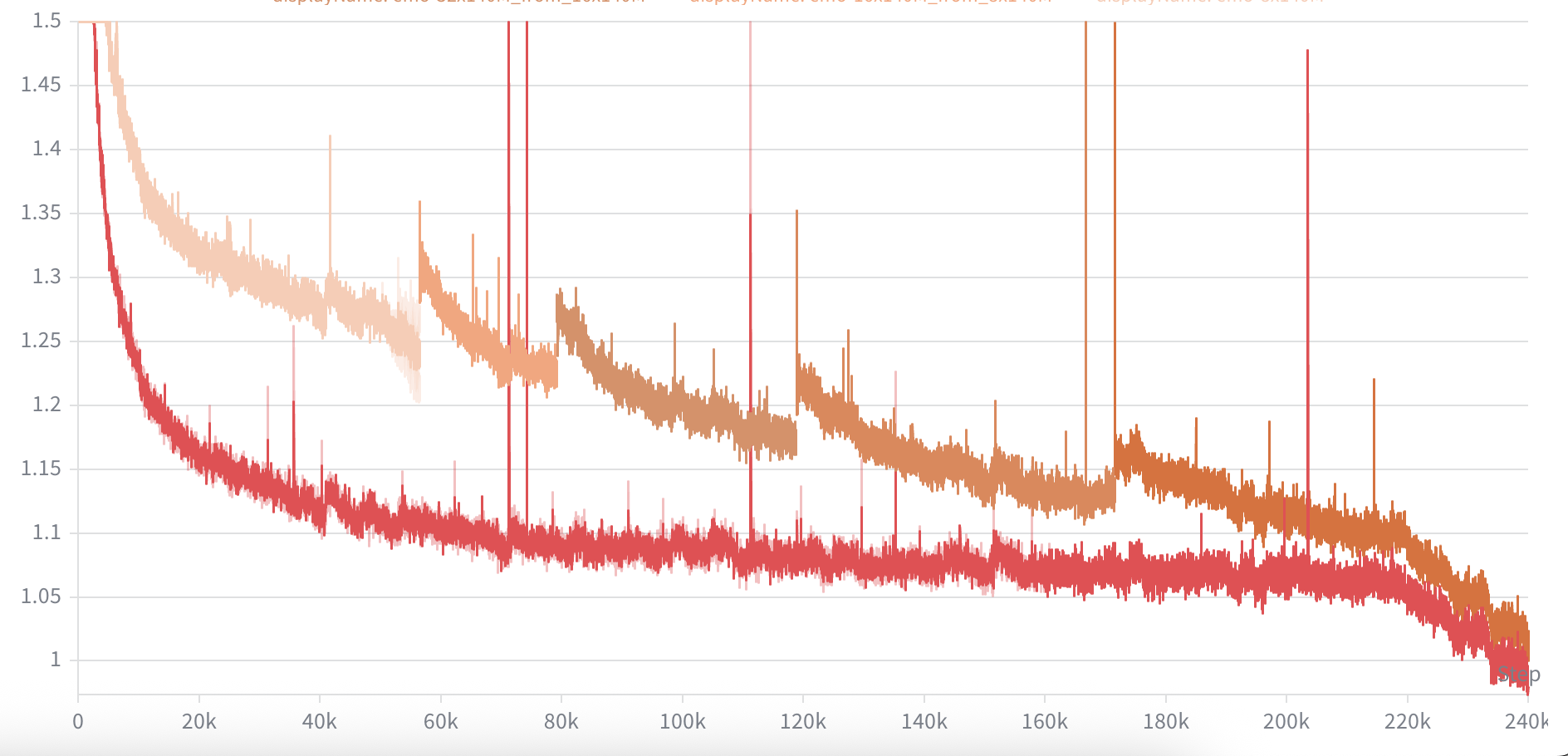}
        \captionof{figure}{Training Loss (not smoothed). Red is Fixed\_E=128, the rest are EMO 5 stages.}
        \label{fig:emo_real_loss}
    \end{minipage}
    \hfill
    \begin{minipage}{0.48\linewidth}
        \centering
        \includegraphics[width=\linewidth]{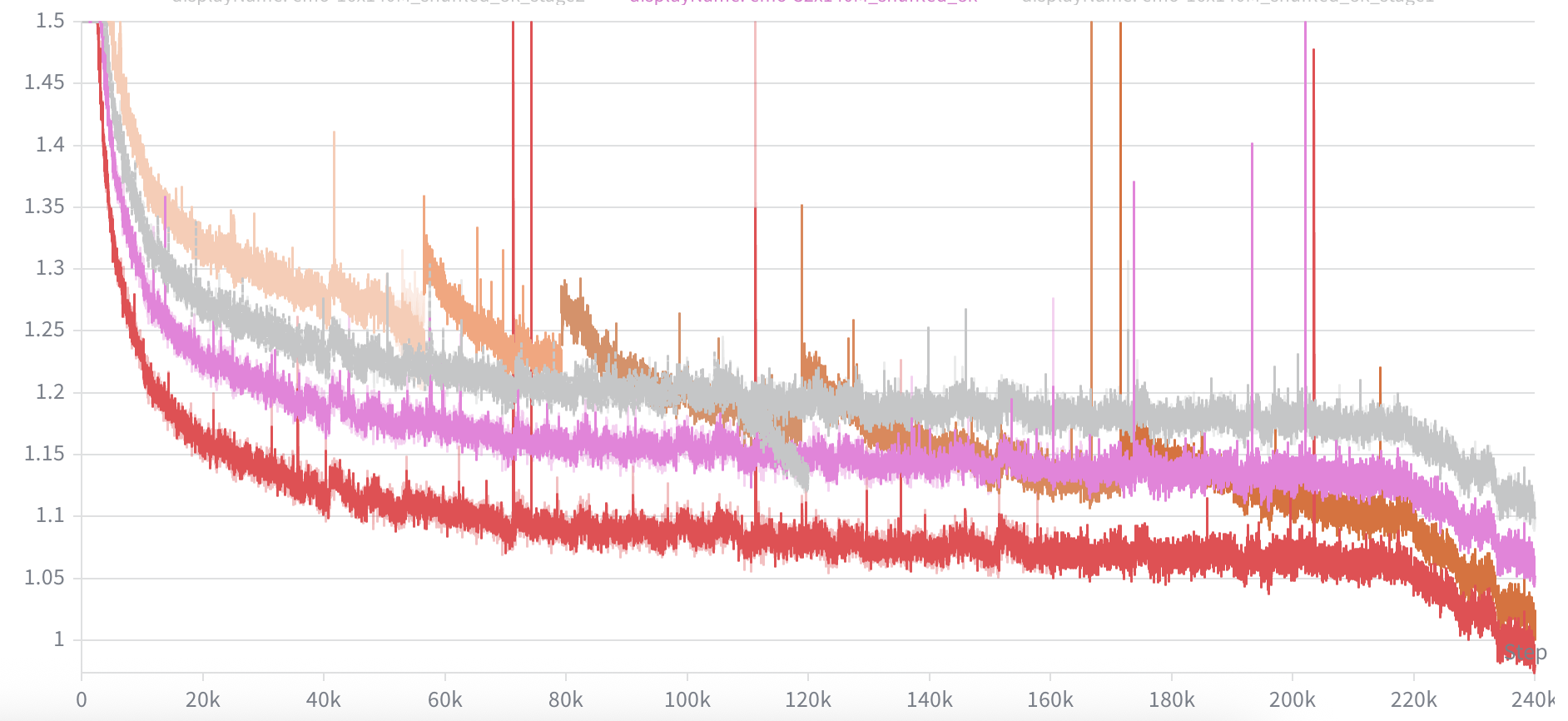}
        \captionof{figure}{Training Loss (not smoothed) with all baselines. Gray is Fixed\_E=16, pink is Fixed\_E=32, red is Fixed\_E=128.}
        \label{fig:emo_real_loss_all}
    \end{minipage}
    \vspace{-10pt}
\end{figure}

Below we show the un-smoothed training loss of main experiments in \autoref{fig:emo_real_loss} and \autoref{fig:emo_real_loss_all}.

\subsection{Validation PPL}
\label{app:val}
\begin{figure}[h]
    \centering
    \includegraphics[width=0.8\linewidth]{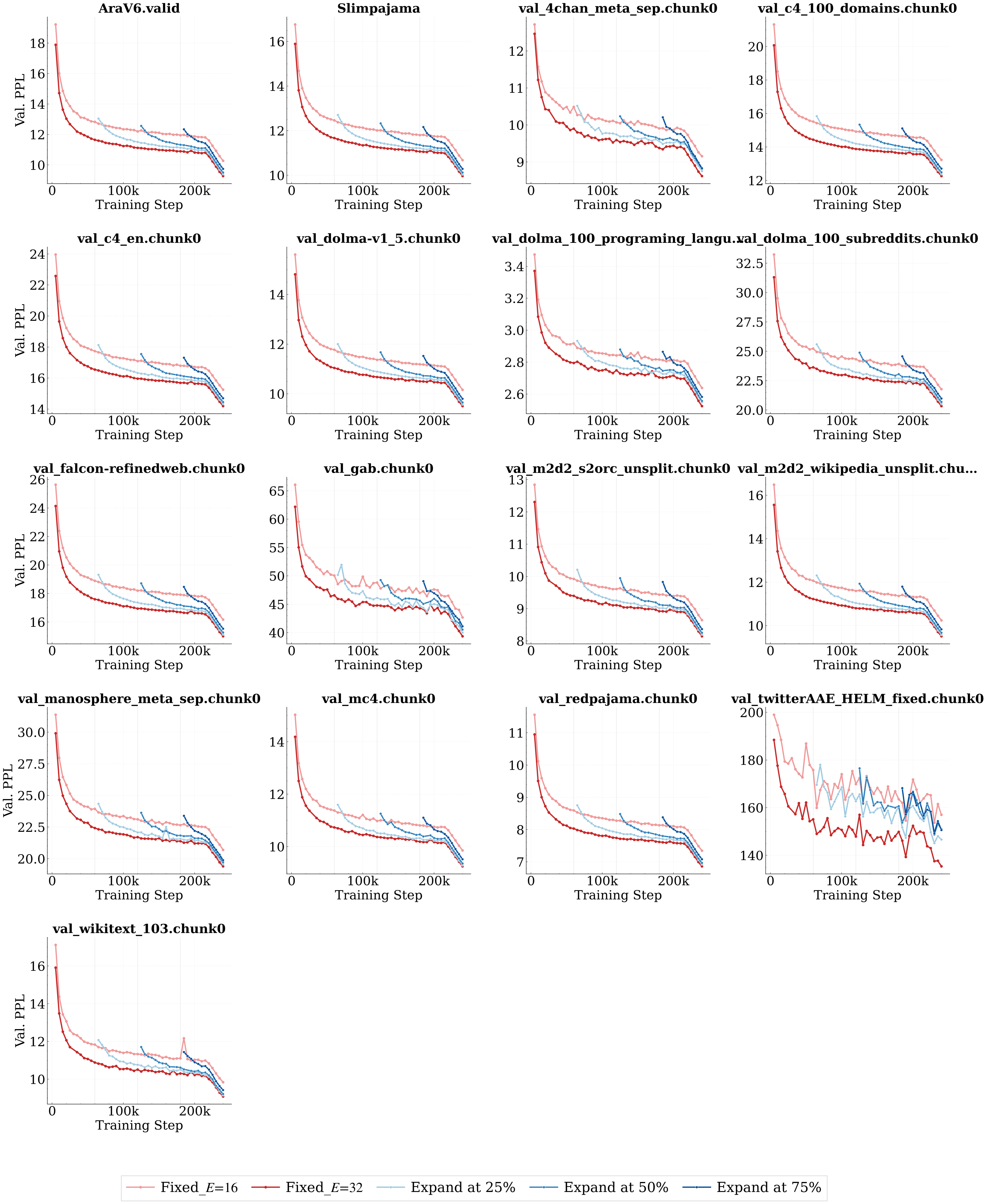}
    \caption{Validation Perplexity of expansion timing experiments (Expand@25\%,50\%, 75\%). Baselines are Fixed\_E=16 and Fixed\_E=32.}
    \label{fig:prelim_val_ppl}
    \vspace{-10pt}
\end{figure}

Figure~\ref{fig:prelim_val_ppl} reports validation perplexity for the preliminary $E=16\rightarrow32$ expansion study used to validate the scaling-law allocation.  This experiment isolates a single expansion boundary and compares different expansion timings.  The key observation is expanding at 25\% and 50\% reaches almost the same perplexity as Fixed\_E=32 baselines in most domains. 

\begin{figure}[h]
    \centering
    \includegraphics[width=1.0\linewidth]{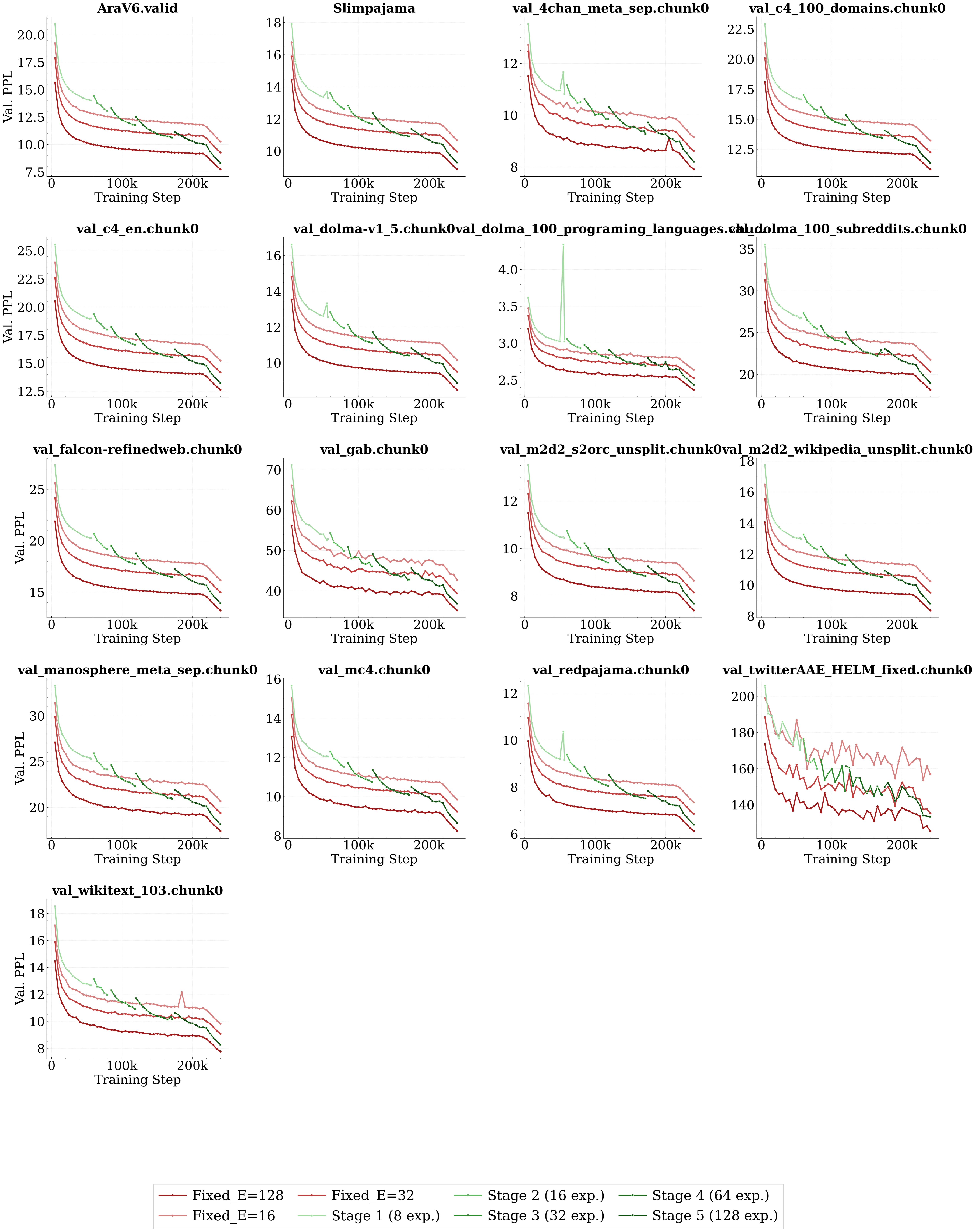}
    \caption{Validation Perplexity of main experiments. Green lines are our progressive training ppls, red lines are baselines from Fixed\_E=16, Fixed\_E=32 and Fixed\_E=128.}
    \label{fig:core_va_ppl}
\end{figure}

Figure~\ref{fig:core_va_ppl} reports the validation perplexity for our main expansion experiments, which validates that in diverse domains of held-out validation sets, EMO is able to achieve comparable performance as fixed baselines.

\section{MOE Scaling Law}

\citet{hoffmann2022training} adapts scaling laws to MoE by expressing loss as a function of activated model size $N_{\text{act}}$ and dataset size $D$:
\begin{equation*}
    L(N_{\text{act}}, D) = m N_{\text{act}}^{\mu} + n D^{\nu} + c,
    \label{eq1}
\end{equation*}

\cite{clark2022unified} studies scaling under fixed datasets while varying both model size and expert count:
\begin{equation*}
    L(N_{\text{act}}, E) = a \hat{E}^{\delta} N_{\text{act}}^{\alpha + \gamma \ln \hat{E}},
    \label{eq2}
\end{equation*}
$\hat{E}$ is a monotonic transformation of the number of experts as defined:
\begin{equation*} 
    \frac{1}{\hat{E}} = \frac{1}{E - 1 + \left(\frac{1}{E_{\text{start}}} - \frac{1}{E_{\text{max}}}\right)^{-1}} + \frac{1}{E_{\text{max}}}.
\end{equation*}

Then, combining with the joint scaling law \cite{ludziejewski2025joint}, we models the loss that depend on activated parameters, total experts, and dataset size:
\begin{equation*}
    L(N_{\text{act}}, E, D) = m(\hat{E}) N_{\text{act}}^{\mu(\hat{E})} + n(\hat{E}) D^{\nu(\hat{E})} + c,
\end{equation*}

To derive the optimal $D$ given a fixed compute budget F and fixed expert count $E$, we need to solve:
\[
\arg\min_{N_{\mathrm{act}}, D} L_{E}(N_{\mathrm{act}}, D)
\quad \text{s.t.} \quad
F = 6 N_{\mathrm{act}} D.
\]

To solve for $D$, substitute:
\[
N_{\mathrm{act}} = \frac{F}{6D},
\]
and set the derivative to $0$:
\[
\frac{dL}{dD}
=
\frac{d}{dD}
\left[
m(\hat{E})\left(\frac{F}{6D}\right)^{\mu(\hat{E})}
+
n(\hat{E}) D^{\nu(\hat{E})}
\right]
= 0.
\]

After rearranging:
\[
D^*(F, E)
=
\left(
\frac{n(\hat{E})\,\nu(\hat{E})}
     {m(\hat{E})\,\mu(\hat{E})}
\right)^{-\frac{1}{\mu(\hat{E})+\nu(\hat{E})}}
\left(\frac{F}{6}\right)^{\frac{\mu(\hat{E})}{\mu(\hat{E})+\nu(\hat{E})}}.
\]

As in our case, we fix model activated parameter $N_{\mathrm{act}}=\bar N_{\mathrm{act}}$, we search over compute-optimal solutions (parameterized by F), and select the one whose $N_{\mathrm{act}}^*$ matches our prescribed $\bar N_{\mathrm{act}}$.

\end{document}